\documentclass[11pt]{article}
\usepackage{amsmath}
\usepackage{amsfonts}
\usepackage{amssymb}
\usepackage{amsthm}

\usepackage{times}
\usepackage{graphicx} 
\usepackage{subfigure} 

\usepackage[authoryear,round]{natbib}

\usepackage{hyperref}

\usepackage{xr}
\newtheorem{thm}{Theorem}
\newtheorem{col}{Corollary}

\newcommand*{\QEDB}{\hfill\ensuremath{\square}}

\title{Convex Coupled Matrix and Tensor Completion}
\author{
Kishan Wimalawarne \\
Bioinformatics Center,\\ 
Institute for Chemical Research,\\
Kyoto University, \\
Gokasho, Uji, Japan.\\
\texttt{kishanwn@gmail.com}
\and Makoto Yamada\\
RIKEN, Center for Advanced Intelligence Project,\\ 
Nihonbashi 1-chome Mitsui Building, 15th floor, \\
1-4-1 Nihonbashi, Chuo-ku, \\ 
Tokyo 103-0027, Japan.\\
The Institute of Statistical Mathematics,\\
10-3 Midori-cho, Tachikawa, \\
Tokyo 190-8562, Japan.\\
PRESTO, Japan Science and Technological Agency (JST), Japan.\\
\texttt{makoto.yamada@riken.jp}
\and Hiroshi Mamitsuka\\
Bioinformatics Center,\\ 
Institute for Chemical Research,\\
Kyoto University, \\
Gokasho, Uji, Japan.\\
Department of Computer Science,\\
Aalto University,\\
Espoo 02150 Finland.\\
\texttt{mami@kuicr.kyoto-u.ac.jp} 
}

\date{}
\begin{document} 
\maketitle
\begin{abstract} 
 We propose a set of convex low-rank inducing norms for  \emph{coupled} matrices and tensors (hereafter coupled tensors), in which  information is shared between the matrices and tensors  through common modes. More specifically, we first propose a mixture of the overlapped trace norm and the latent norms with the matrix trace norm, and then, propose a completion model regularized using these norms to impute coupled tensors.  A key advantage of the proposed norms is that they are convex and can be used to find a globally optimal solution, whereas existing methods for coupled learning are non-convex. We also analyze the excess risk bounds of the completion model regularized using our proposed norms and show that they can exploit the low-rankness of coupled tensors, leading to better bounds compared to those obtained using uncoupled norms. Through synthetic and real-data experiments, we show that the proposed completion model compares favorably with existing ones.
\end{abstract}

\section{Introduction}
Learning from a matrix or a tensor has long been an important problem in machine learning. In particular,  matrix and tensor factorization using low-rank inducing norms has been studied extensively, and  many applications have been considered, such as  missing value imputation \citep{SigDinDeLSuy13, DBLP:conf/iccv/LiuMWY09}, multi-task learning \citep{DBLP:conf/nips/ArgyriouEP06, DBLP:conf/icml/Romera-ParedesABP13, nips-14}, subspace clustering \citep{conf/icml/LiuLY10},  and inductive learning \citep{SigDinDeLSuy13,DBLP:journals/neco/WimalawarneTS16}. 
 Though useful in many applications, factorization based on an individual matrix or tensor tends to perform poorly under the \emph{cold start} setup condition \citep{singh2008relational}, when, for example, it is not possible to observe  click information for new users in collaborative filtering. It therefore cannot be used to recommend possible items for new users. Potential ways to address this issue, are matrix or tensor factorization with  side information \citep{Narita2011}. Both have been applied to recommendation systems \citep{singh2008relational,gunasekar2015consistent} and personalized medicine \citep{Khan2014}.

Both matrix and tensor factorization with side information can be regarded as the joint factorization of  coupled matrices and tensors (hereafter coupled tensors) (See Figure~\ref{fig:1}).
 \citet{journals/corr/abs-1105-3422} introduced a coupled factorization method based on CP decomposition, that simultaneously factorizes matrices and tensors by sharing the low-rank structures in the matrices and tensors. The coupled factorization approach has been applied to joint analysis of fluorescence and proton nuclear magnetic resonance (NMR) measurements \citep{DBLP:conf/eusipco/AcarNS14} and joint  NMR  and liquid chromatography-mass spectrometry (LC–MS)  \citep{7202834}. More recently, a Bayesian approach  proposed by \citet{journals/datamine/ErmisAC15}  was applied to link prediction problems. However, existing \emph{coupled} factorization methods are \emph{non-convex} and can obtain only a poor local optimum. Moreover, the ranks of the coupled tensors need to be determined beforehand. In practice, it is difficult to specify the true ranks of the tensor and the matrix without prior knowledge. Furthermore, existing algorithms are not theoretically guaranteed.

We propose in this paper,  \emph{convex} norms for coupled tensors that overcome the non-convexity problem. The norms are a mixtures of tensor norms: the overlapped trace norm \citep{conf/nips/TomiokaSHK11}, the latent trace norm \citep{tomioka/nips13/abs-1303-6370}, the  scaled latent norm \citep{nips-14}, and the matrix trace norm \citep{DBLP:conf/nips/ArgyriouEP06}.  A key advantage of the proposed norms is that they are convex and thus can be used to find a globally optimal solution, whereas existing  coupled factorization approaches are non-convex. Furthermore, we analyze the excess risk bounds of the completion model regularized using our proposed norms. Through synthetic and real-data experiments, we show that it  compares favorably with existing ones.

Our contributions in this paper are to
\vspace{.05in}
\begin{itemize}
\item Propose  a set of convex coupled norms for matrices and tensors that extend low-rank tensor and matrix norms.
\item Propose mixed norms that combine features from both the overlapped norm and latent norms.
\item Propose a convex completion model  regularized using the proposed coupled norms. 
\item Analyze the excess risk bounds for the proposed completion model with respect to the proposed norms and show that it leads to lower excess risk. 
\item Show through synthetic and real-data experiments, that our norms lead to  performance comparable to that of  existing non-convex methods. 
\item Show that our norms are applicable to coupled tensors  based on both the CP rank and the multilinear rank without prior assumptions about their low-rankness.
\item Show that convexity  of the proposed norms leads to global solutions, eliminating the need to deal with local optimal solutions as is necessary with non-convex methods. 
\end{itemize}

The remainder of the paper is organized as follows.
In Section~2, we discuss  related work on  coupled tensor completion. 
In Section~3, we present our proposed method, first introducing a coupled completion model and then proposing a  set of norms called  coupled norms.
In Section~4, we give optimization methods for solving the coupled completion model.
In Section~5, we  theoretically analyze it using excess risk bounds for the  proposed coupled norms.
In Section~6, we present the results of our evaluation using synthetic  and real-world data experiments.
Finally, in Section~7, we summarize the key points and suggest future work. 

\section{Related Work}
Most of the models that  proposed for learning with multiple matrices or tensors use joint factorization of matrices and tensors. The regularization-based model proposed by \citet{journals/corr/abs-1105-3422} for completion of coupled tensors and which was further studied  \citep{DBLP:conf/eusipco/AcarNS14,DBLP:journals/bmcbi/AcarPGRLNB14,7202834} uses  CANDECOMP/PARAFAC (CP) decomposition \citep{Carroll1970,harshman-parafac-1970,hitchcock-sum-1927,journals/siamrev/KoldaB09} to factorize the tensor and operates under the assumption  that the factorized components of its coupled mode are in common with the  factorized components of the matrix on the same mode.  Bayesian models have also been proposed for imputing missing values  with applications in link prediction \citep{journals/datamine/ErmisAC15} and non-negative factorization \citep{6729621} which  use similar factorization models. Applications that have used collective factorization of tensors are multi-view factorization \citep{Khan2014} and multi-way clustering \citep{doi:10.1137/1.9781611972771.14}. Due to their use of factorization-based learning, all of these models are \emph{non-convex}. 

The use of common adjacency graphs  has more recently been proposed for  incorporating  similarities among heterogeneous tensor data \citep{Li:2015:MCC:2886521.2886703}. Though this method does not require assumptions about rank for explicit factorization of tensors, it depends on the modeling of the common adjacency graph and does not incorporate the low-rankness created by the coupling of tensors.

\section{Proposed Method}
\begin{figure}[h]
\centering
\includegraphics[scale=0.6]{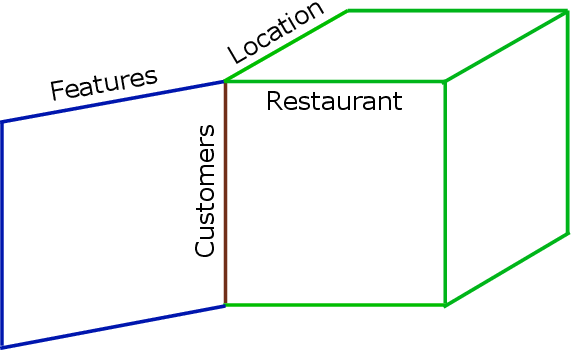}
\caption{Illustration of information sharing between matrix and tensor in coupled tensor,  through "Customers" mode.}
\label{fig:1}
\end{figure}

We investigate a method for coupling  a matrix and a tensor that  forms when they share a common mode \citep{7202834,DBLP:conf/eusipco/AcarNS14,DBLP:journals/bmcbi/AcarPGRLNB14}. An example of the most basic coupling is shown in Figure \ref{fig:1} where a $3$-way (third-order) tensor is attached to a matrix on a specific mode. As depicted, we may have a problem of predicting recommendations for customers  on the basis of their preferences of restaurants in different locations and we may also have side information about the characteristics for each customer. We can utilize this side information, by coupling the customer-characteristic matrix with the sparse customer-restaurant-location tensor of the customer mode and then  impute the missing values in the tensor.    

Let us consider a partially observed  matrix $\hat{M} \in \mathbb{R}^{n_{1} \times m}$ and a partially observed $3$-way tensor $\hat{\mathcal{T}} \in \mathbb{R}^{n_{1} \times n_{2} \times n_{3}}$ with mappings to observed elements indexed by $\Omega_{M}$ and $\Omega_{\mathcal{T}}$, respectively, and let us assume that they are  coupled on the first mode. Our ultimate goal of this paper is to introduce  convex coupled norms  $\|  \mathcal{T} , M \|_{\mathrm{cn}}$ for use in solving 
\begin{equation}
\underset{\mathcal{T} , M  }{\min} \frac{1}{2}\| \Omega_{M}(M - \hat{M}) \|_{\mathrm{F}}^{2} + \frac{1}{2}\| \Omega_{\mathcal{T}}(\mathcal{T} - \hat{\mathcal{T}}) \|_{\mathrm{F}}^{2} + \lambda \|  \mathcal{T} , M \|_{\mathrm{cn}} \label{eq:coup_objective}, 
\end{equation} 
where  $\lambda \geq 0$ is the regularization parameter. We also investigate the theoretical properties of problem \eqref{eq:coup_objective}.

\textbf{Notations:} The mode-$k$ unfolding of  tensor $\mathcal{T} \in \mathbb{R}^{n_{1} \times \cdots \times n_{K}}$  is represented as $T_{(k)} \in \mathbb{R}^{n_{k}\times \prod_{j \neq k}^{K}n_{j}}$, which is obtained by concatenating all the $\prod_{j \neq k}^{K}n_{j}$ vectors with dimension  $n_{k}$ obtained by fixing all except the $k$th index on  mode-$k$ along its columns.  We use  $vec()$ to indicate the conversion of a matrix or a tensor into a vector and $unvec()$ to represent the reverse operation. The spectral norm (operator norm) of a matrix $X$ is the $\|X\|_{{\mathrm{op}}} $ that is the largest singular value of $X$. The Frobenius norm of a tensor $\mathcal{T}$ is defined as $\|\mathcal{T}\|_{{\mathrm{F}}} = \sqrt{ \left \langle \mathcal{T} , \mathcal{T} \right \rangle } = \sqrt{vec(\mathcal{T})^{\top}vec(\mathcal{T})}$. We use  $[M;N]$ as the concatenation of matrices $M \in \mathbb{R}^{m_{1} \times m_{2}}$ and $N \in \mathbb{R}^{m_{1} \times m_{3}}$ along their mode $1$.

\subsection{Existing Matrix and Tensor Norms} 

Before we introduce our new norms, let us first briefly review the existing low-rank inducing matrix and tensor norms. Among matrices, the matrix trace norm \citep{DBLP:conf/nips/ArgyriouEP06}  is a commonly used convex relaxation for the minimization of the  rank of a matrix. For a given matrix $M \in \mathbb{R}^{n_{1} \times m}$ with rank $J$, we can define its trace norm as 
\begin{equation*}
\|M\|_{{\mathrm{tr}}} = \sum_{j=1}^{J} \sigma_{j}, \label{eq:matrix_trace}
\end{equation*}
where $\sigma_{j}$ is the $j$th non-zero singular value of the matrix.

Low-rank inducing norms for tensors have received revived attention  in  recent years. One of the earliest low-rank inducing tensor norm is the  \emph{tensor nuclear norm} \citep{DBLP:conf/iccv/LiuMWY09} (also known as the \emph{overlapped trace norm} \citep{tomioka/nips13/abs-1303-6370}) which can be expressed for a tensor
$\mathcal{T} \in \mathbb{R}^{n_{1}\times \cdots \times n_{K}}$ as 
\begin{equation}
\| \mathcal{T} \|_{\mathrm{overlap}} = \underset{k=1}{\overset{K}{\sum}} \|T_{(k)}\|_{\mathrm{tr}}.\label{eq:overlap_norm}
\end{equation}

\citet{tomioka/nips13/abs-1303-6370} proposed   the  \emph{latent trace norm}:
\begin{equation}
\| \mathcal{T} \|_{\mathrm{latent}} = \underset{\mathcal{T}^{(1)} + \ldots +\mathcal{T}^{(K)} = \mathcal{T} }{\inf} \underset{k=1}{\overset{K}{\sum}}  \| T_{(k)}^{(k)} \|_{\mathrm{tr}}.\label{eq:latent_norm}
\end{equation}

The \emph{scaled latent trace norm} was proposed as an extension of the latent trace norm \citep{nips-14}:  
\begin{equation}
\| \mathcal{T} \|_{\mathrm{scaled}} = \underset{\mathcal{T}^{(1)} + \ldots +\mathcal{T}^{(K)} = \mathcal{T} }{\inf} \underset{k=1}{\overset{K}{\sum}} \frac{1}{\sqrt{n_{k}}} \| T_{(k)}^{(k)} \|_{\mathrm{tr}}. \label{eq:scaled_norm}
\end{equation}

The behaviors of these two tensor norms have been studied on the basis of  multitask learning  \citep{nips-14}  and inductive learning  \citep{DBLP:journals/neco/WimalawarneTS16}. The results show that for a tensor $\mathcal{T} \in \mathbb{R}^{n_{1} \times \cdots \times n_{K}}$ with multilinear rank $(r_{1},\ldots,r_{K})$, the excess  risk is  bounded above with respect to regularization with the overlapped trace norm  by $\mathcal{O}(\sum_{k=1}^{K} \sqrt{r_{k}})$, the latent trace norm  by $\mathcal{O}(\min_{k} \sqrt{r_{k}})$, and  the scaled latent trace norm  by $\mathcal{O}\Big(\min_{k} \sqrt{\frac{r_{k}}{n_{k}}}\Big)$.

\subsection{Coupled Tensor Norms}
As with individual matrices and tensors, having  convex and low-rank inducing norms for coupled tensors would be useful in achieving global solutions for coupled tensor completion with theoretical guarantees. To achieve this, we propose a set of  norms for coupled tensors that are coupled on specific modes using  existing matrix and tensor trace norms. Let us first  define a new coupled norm with the format  $\| . \|^{a}_{(b,c,d)}$, where the superscript $a$ specifies the mode in which the tensor and  matrix are coupled and the subscripts  $b,c,d \in \{\mathrm{O},\mathrm{L},\mathrm{S},-\}$ indicate how the modes are regularized. 
The notations for $b,c,d$ are defined as
\begin{description}  
\item  $\mathrm{O}$: The mode is regularized with the trace norm. The same tensor is regularized on other modes  similarly to the overlapped trace norm. 
\item  $\mathrm{L}$: The mode is  considered to be a  latent tensor  that is regularized using the trace norm only with respect to that mode.
\item  $\mathrm{S}$: The mode is regularized as a latent tensor but it is  scaled similarly to the scaled latent trace norm.
\item  $-$: The mode is not regularized.
\end{description}

Given a matrix $M \in \mathbb{R}^{n_{1} \times m}$ and a  tensor $\mathcal{T} \in \mathbb{R}^{n_{1} \times n_{2} \times n_{3}}$, we introduce three norms that are  \emph{coupled} extensions of the overlapped trace norm, the latent trace norm, and the scaled latent trace norm, respectively.

\vspace{.1in}
\noindent {\bf Coupled overlapped trace norm:}
\begin{equation}
\| \mathcal{T} , M  \|^{1}_{(\mathrm{O},\mathrm{O},\mathrm{O})} :=   \| [T_{(1)};M] \|_{\mathrm{tr}} + \sum_{k=2}^{3} \|T_{(k)} \|_{\mathrm{tr}}.  \label{eq:coup_overlap}
\end{equation}

\vspace{.1in}
\noindent {\bf Coupled latent trace norm:}
\begin{equation}
\| \mathcal{T} , M  \|^{1}_{(\mathrm{L},\mathrm{L},\mathrm{L})} =  \underset{\mathcal{T}^{(1)} + \mathcal{T}^{(2)} +  \mathcal{T}^{(3)} = \mathcal{T} }{\inf} \bigg( \| [T_{(1)}^{(1)};M] \|_{\mathrm{tr}} 
+ \sum_{k=2}^{3} \|T_{(k)}^{(k)} \|_{\mathrm{tr}} \bigg). \label{eq:coup_latent}
\end{equation}

\vspace{.1in}
\noindent {\bf Coupled scaled latent trace norm:}
\begin{equation}
\| \mathcal{T} , M  \|^{1}_{(\mathrm{S},\mathrm{S},\mathrm{S})} =    \underset{\mathcal{T}^{(1)} + \mathcal{T}^{(2)} + \mathcal{T}^{(3)} = \mathcal{T} }{\inf} \bigg( \frac{1}{\sqrt{n_{1}}} \| [T_{(1)}^{(1)};M] \|_{\mathrm{tr}}
 + \sum_{k=2}^{3} \frac{1}{\sqrt{n_{k}}} \|T_{(k)}^{(k)} \|_{\mathrm{tr}} \bigg). \label{eq:coup_scaled}
\end{equation}

In addition to these norms, we can also create norms as mixtures of overlapped and latent/scaled latent norms. For example, if we want to create a norm that is regularized using the scaled latent trace norm on the second mode while the other modes are regularized using the overlapped trace norm, we can define it as

\vspace{.1in}
\begin{multline}  
\| \mathcal{T} , M  \|^{1}_{(\mathrm{O},\mathrm{S},\mathrm{O})} =  \underset{\mathcal{T}^{(1)} + \mathcal{T}^{(2)} = \mathcal{T} }{\inf} \bigg( \| [T_{(1)}^{(1)};M] \|_{\mathrm{tr}} 
 +  \frac{1}{\sqrt{n_{2}}} \|T_{(2)}^{(2)} \|_{\mathrm{tr}}  \\
 +  \|T_{(3)}^{(1)} \|_{\mathrm{tr}} \bigg). \label{eq:mixture_020}
\end{multline} 
This norm has two latent tensors, $\mathcal{T}^{(1)}$ and $\mathcal{T}^{(2)}$. Tensor $\mathcal{T}^{(1)}$ is regularized using the overlapped method for modes $1$ and $3$ while the  tensor $\mathcal{T}^{(2)}$ is regularized as a scaled latent tensor on mode $2$. Given this use of a mixture of regularization methods, we call the resulting norm a \emph{mixed norm}.

In a similar manner, we can create other mixed norms distinguished by their subscripts: $(\mathrm{L},\mathrm{O},\mathrm{O})$, $(\mathrm{O},\mathrm{L},\mathrm{O})$, $(\mathrm{O},\mathrm{O},\mathrm{L})$, $(\mathrm{S},\mathrm{O},\mathrm{O})$, $(\mathrm{O},\mathrm{S},\mathrm{O})$, and  $(\mathrm{O},\mathrm{O},\mathrm{S})$. The main advantage gained by using these mixed norms is  the additional freedom to regularize low-rank constraints among coupled tensors. Other combinations of norms in which two modes are latent  tensors such as $(\mathrm{L},\mathrm{L},\mathrm{O})$ will make the third mode also a latent tensor since overlapped regularization requires that more than one mode be regularized of the same tensor. Though we have considered using the latent trace norm, in practice it has been shown to be weaker in performance than the scaled latent trace norm \citep{nips-14,DBLP:journals/neco/WimalawarneTS16}. Therefore, in our experiments, we  considered only mixed norms based on the scaled latent trace norm. 

\subsubsection{Extensions for Multiple Matrices and Tensors}
Our newly defined norms can be extended to multiple matrices coupled to a tensor on different modes. For instance, we can couple two matrices $M_{1} \in \mathbb{R}^{n_{1} \times m_{1}}$ and $M_{2} \in \mathbb{R}^{n_{3} \times m_{2}}$ to a $3$-way tensor $\mathcal{T}$ on its first and third modes. If we regularize the coupled tensor with the overlapped trace norm on mode $1$ and mode $3$ and the scaled latent trace norm on mode $2$, we  obtain a mixed norm, 
\begin{multline*}  
\| \mathcal{T} , M_{1}, M_{2} \|^{1,3}_{(\mathrm{O},\mathrm{S},\mathrm{O})} =  \underset{\mathcal{T}^{(1)} + \mathcal{T}^{(2)} = \mathcal{T} }{\inf} \Big( \| [T_{(1)}^{(1)};M_{1}] \|_{\mathrm{tr}}
 +  \frac{1}{\sqrt{n_{2}}} \|T_{(2)}^{(2)} \|_{\mathrm{tr}} \\
 +  \| [T_{(3)}^{(1)};M_{2}] \|_{\mathrm{tr}} \Big). 
\end{multline*}

Coupled norms for multiple $3$-mode or higher dimensional tensors could also be designed using our proposed method. However, such extension may require extending coupled norms further. Extensions to coupled norms for multiple tensors are a promising area for  future research.

\subsection{Dual Norms}
Let us now briefly look at   dual norms for the above defined coupled norms. Dual norms are useful in deriving excess risk bounds, as discussed in Section 4. Due to space limitations we  derive  dual norms for only two coupled norms  to better understand their nature. To derive them, we first need to know the Schatten norm \citep{tomioka/nips13/abs-1303-6370} for the coupled tensor norms. Let us first define the Schatten-$(p,q)$ norm  for the coupled  norm $\| \mathcal{T} , M  \|^{1}_{(\mathrm{O},\mathrm{O},\mathrm{O})}$  with an additional subscript notation $\underline{S_{p}/q}$:
\begin{multline}
\| \mathcal{T} , M  \|_{(\mathrm{O},\mathrm{O},\mathrm{O}),\underline{S_{p}/q} }^{1} := \Bigg(  \bigg( \sum_{i}^{r_{1}} \sigma_{i}\big( [T_{(1)};M]\big)^{p} \bigg)^{\frac{q}{p}} 
 +  \bigg( \sum_{j}^{r_{2}} \sigma_{j}\big(T_{(2)} \big)^{p} \bigg)^{\frac{q}{p}} \\
  + \bigg( \sum_{k}^{r_{3}} \sigma_{k}\big(T_{(3)} \big)^{p} \bigg)^{\frac{q}{p}} \Bigg)^{\frac{1}{q}}, \label{eq:sp_000}
\end{multline}
where $p$ and $q$ are constants, $r_{1}$, $r_{2}$ and $r_{3}$ are the ranks and $\sigma_{i}$, $\sigma_{j}$  and $\sigma_{k}$ are the singular values for each unfolding.

The following theorem presents the dual norm of $\| \mathcal{T} , M  \|_{(\mathrm{O},\mathrm{O},\mathrm{O}),\underline{S_{p}/q}}^{1}$ (see  Appendix A for proof). 

\begin{thm}
Let a matrix  $M \in \mathbb{R}^{n_{1} \times m}$ and a tensor $\mathcal{T} \in \mathbb{R}^{n_{1} \times n_{2} \times n_{3}}$ be coupled on their first modes. The dual norm of    $\| \mathcal{T} , M  \|_{(\mathrm{O},\mathrm{O},\mathrm{O}),\underline{S_{p}/q} }^{1}$ with $1/p + 1/p^{*} =1$ and $1/q + 1/q^{*} =1$ is 
\begin{multline*}
\| \mathcal{T} , M  \|_{(\mathrm{O},\mathrm{O},\mathrm{O}),{\overline{S_{p^{*}}/q^{*}}}}^{1} = 
 \underset{\mathcal{T}^{(1)} + \mathcal{T}^{(2)} +  \mathcal{T}^{(3)} = \mathcal{T} }{\inf} \Bigg(  \bigg( \sum_{i}^{r_{1}} \sigma_{i}\big( [T_{(1)}^{(1)};M]\big)^{p^{*}} \bigg)^{\frac{q^{*}}{p^{*}}} \\
 + \bigg( \sum_{j}^{r_{2}} \sigma_{j}\big(T_{(2)}^{(2)} \big)^{p^{*}} \bigg)^{\frac{q^{*}}{p^{*}}}
 + \bigg( \sum_{k}^{r_{3}} \sigma_{k}\big(T_{(3)}^{(3)} \big)^{p^{*}} \bigg)^{\frac{q^{*}}{p^{*}}} \Bigg)^{\frac{1}{q^{*}}}, 
\end{multline*}
where $r_{1}$, $r_{2}$, and $r_{3}$ are the ranks for each mode and $\sigma_{i}$, $\sigma_{j}$,  and $\sigma_{k}$ are the singular values for each unfolding of the coupled tensor.
\end{thm}

In the special case of $p = 1$ and $q = 1$, we see that $\| \mathcal{T} , M  \|_{(\mathrm{O},\mathrm{O},\mathrm{O}),\underline{S_{1}/1} }^{1} = \| \mathcal{T} , M  \|_{(\mathrm{O},\mathrm{O},\mathrm{O})}^{1}$. Its dual norm is the spectral norm, as shown in the following corollary.

\begin{col}
Let a matrix  $M \in \mathbb{R}^{n_{1} \times m}$ and a tensor $\mathcal{T} \in \mathbb{R}^{n_{1} \times n_{2} \times n_{3}}$ be coupled on their first mode. The dual norm of    $\| \mathcal{T} , M  \|_{(\mathrm{O},\mathrm{O},\mathrm{O}),\underline{S_{1}/1} }^{1}$ is 
\begin{multline*}
\| \mathcal{T} , M  \|_{(\mathrm{O},\mathrm{O},\mathrm{O}),{\overline{S_{\infty}/\infty}}}^{1} = \\
 \underset{\mathcal{T}^{(1)} + \mathcal{T}^{(2)} +  \mathcal{T}^{(3)} = \mathcal{T} }{\inf} \max \Big( \|[T_{(1)}^{(1)};M]\|_{\mathrm{op}},  \|T_{(2)}^{(2)}\|_{\mathrm{op}}, \|T_{(3)}^{(3)}\|_{\mathrm{op}} \Big). 
\end{multline*}
\end{col}

The Schatten-$(p,q)$ norm for the mixed norm $\| \cdot \|^{1}_{(\mathrm{L},\mathrm{O},\mathrm{O})}$ is defined as 
\begin{multline*}
\| \mathcal{T} , M  \|_{(\mathrm{L},\mathrm{O},\mathrm{O}),\underline{S_{p}/q}}^{1} = \underset{\mathcal{T}^{(1)} + \mathcal{T}^{(2)}  = \mathcal{T} }{\inf} \Bigg( \bigg(  \sum_{i}^{r_{1}} \sigma_{i}\big( [T_{(1)}^{(1)};M]\big)^{p} \bigg)^{\frac{q}{p}} \\ 
+ \bigg(  \sum_{j}^{r_{2}} \sigma_{j}\big(T_{(2)}^{(2)} \big)^{p} \bigg)^{\frac{q}{p}}  + \bigg(  \sum_{k}^{r_{3}} \sigma_{k}\big(T_{(3)}^{(2)} \big)^{p} \Big)^{\frac{q}{p}} \Bigg)^{\frac{1}{q}}.
\end{multline*}
Its dual norm is defined by the following theorem (see Appendix A for proof). 

\begin{thm}
Let a matrix  $M \in \mathbb{R}^{n_{1} \times m}$ and a tensor $\mathcal{T} \in \mathbb{R}^{n_{1} \times n_{2} \times n_{3}}$ be coupled on their first mode. The dual norm of the  mixed coupled   norm $\| \mathcal{T} , M  \|_{(\mathrm{L},\mathrm{O},\mathrm{O}),\underline{S_{p}/q} }^{1}$ with $1/p + 1/p^{*} =1$ and $1/q + 1/q^{*} =1$ is 
\begin{multline*}
\| \mathcal{T} , M  \|_{(\mathrm{L},\mathrm{O},\mathrm{O}),\overline{S_{p^{*}}/q^{*}} }^{1} = 
\Bigg( \Big( \sum_{i}^{r_{1}} \sigma_{i}\big( [T_{(1)};M]\big)^{p^{*}} \Big)^{\frac{q^{*}}{p^{*}}} + \\
   \underset{\mathcal{\hat{T}}^{(1)} +  \mathcal{\hat{T}}^{(2)} = \mathcal{T} }{\inf} \Bigg( \bigg( \sum_{j}^{r_{2}} \sigma_{j}\big(\hat{T}_{(2)}^{(1)} \big)^{p^{*}} \bigg)^{\frac{q^{*}}{p^{*}}} 
  + \bigg( \sum_{k}^{r_{3}} \sigma_{k}\big(\hat{T}_{(3)}^{(2)} \big)^{p^{*}} \bigg)^{\frac{q^{*}}{p^{*}}} \Bigg) \Bigg)^{\frac{1}{q^{*}}} , 
\end{multline*}
where $r_{1}$, $r_{2}$, and $r_{3}$ are the ranks of $T_{(1)}$, ${\hat{T}}^{(1)}_{(2)}$ and ${\hat{T}}^{(2)}_{(3)}$, respectively, and $\sigma_{i}$, $\sigma_{j}$, and $\sigma_{k}$ are their singular values.
\end{thm}  
The dual norms of other mixed norms can be similarly derived.

\section{Optimization}
In this section, we discuss optimization of the proposed  completion model  \eqref{eq:coup_objective}. The completion model \eqref{eq:coup_objective} can be easily solved for each coupled norm using a state of the art optimization method such as the alternating direction method of multipliers (ADMM) method \citep{Boyd:2011:DOS:2185815.2185816}. The optimization steps for the coupled norm $\| \mathcal{T} , M  \|^{1}_{(\mathrm{S},\mathrm{O},\mathrm{O})}$ are derived using the ADMM method. The optimization steps for the other norms are similarly  derived. 

We express \eqref{eq:coup_objective} using the  $\| \mathcal{T} , M  \|^{1}_{(\mathrm{S},\mathrm{O},\mathrm{O})}$ norm
\begin{multline}
\underset{\mathcal{T}^{(1)}, \mathcal{T}^{(2)}, M  }{\min} \frac{1}{2}\| \Omega_{M}(M - \hat{M}) \|_{\mathrm{F}}^{2} + \frac{1}{2}\| \Omega_{\mathcal{T}}(\mathcal{T}^{(1)} + \mathcal{T}^{(2)} - \hat{\mathcal{T}}) \|_{\mathrm{F}}^{2} \\
 + \lambda \Bigg(\frac{1}{\sqrt{n_{1}}}  \| [T_{(1)}^{(1)};M] \|_{\mathrm{tr}}  +  \|T_{(2)}^{(2)} \|_{\mathrm{tr}} +  \|T_{(3)}^{(2)} \|_{\mathrm{tr}} \Bigg) \label{eq:coup_objective_overlap}
\end{multline}

By introducing auxiliary variables $X \in \mathbb{R}^{n_{1} \times m}$ and $\mathcal{Y} \in \mathbb{R}^{n_{1} \times n_{2} \times n_{3}}$, we can formulate the objective function of ADMM for \eqref{eq:coup_objective_overlap}  
\begin{multline}
\underset{\mathcal{T}^{(1)}, \mathcal{T}^{(2)}, M  }{\min} \frac{1}{2}\| \Omega_{M}(M - \hat{M}) \|_{\mathrm{F}}^{2} + \frac{1}{2}\| \Omega_{\mathcal{T}}(\mathcal{T}^{(1)} + \mathcal{T}^{(2)} - \hat{\mathcal{T}}) \|_{\mathrm{F}}^{2} \\
 + \lambda \Bigg(\frac{1}{\sqrt{n_{1}}}  \| [Y_{(1)}^{(1)};X] \|_{\mathrm{tr}} 
 +  \|Y_{(2)}^{(2)} \|_{\mathrm{tr}} +  \|Y_{(3)}^{(2)} \|_{\mathrm{tr}} \Bigg) \label{eq:coup_objective_overlap_legrange}
\end{multline} 
\begin{equation*}
{\mathrm{s.t.}} \quad  X = M \quad,\quad \mathcal{Y}^{(1)} = \mathcal{T}^{(1)} ,\quad \mathcal{Y}^{(k)} = \mathcal{T}^{(2)} \quad k = 2,3. 
\end{equation*} 

We introduce Lagrangian multipliers $W^{M} \in  \mathbb{R}^{n_{1} \times m}$  and $\mathcal{W}^{\mathcal{T}^{(k)}} \in \mathbb{R}^{n_{1} \times n_{2} \times n_{3}}$, $(k =1,2,3)$ and formulate the Lagrangian as 
\begin{multline}
\underset{\mathcal{T}^{(1)}, \mathcal{T}^{(2)}, M  }{\min} \frac{1}{2}\| \Omega_{M}(M - \hat{M}) \|_{\mathrm{F}}^{2} + \frac{1}{2}\| \Omega_{\mathcal{T}}(\mathcal{T}^{(1)} + \mathcal{T}^{(2)} - \hat{\mathcal{T}}) \|_{\mathrm{F}}^{2} \\
 + \lambda \Bigg(\frac{1}{\sqrt{n_{1}}}  \| [Y_{(1)}^{(1)};X] \|_{\mathrm{tr}} 
 +  \|Y_{(2)}^{(2)} \|_{\mathrm{tr}} +  \|Y_{(3)}^{(2)} \|_{\mathrm{tr}} \Bigg) + \big\langle W^{M}, M -X \big\rangle \\
  + \big\langle \mathcal{W}^{\mathcal{T}^{(1)}}, \mathcal{T}^{(1)} - \mathcal{Y}^{(1)} \big\rangle 
 + \sum_{k=2}^{3} \big\langle \mathcal{W}^{\mathcal{T}^{(k)}}, \mathcal{T}^{(2)} - \mathcal{Y}^{(k)} \big\rangle  + \frac{\beta}{2}\| M -X \|_{F}^{2} 
\\
  + \frac{\beta}{2} \| \mathcal{T}^{(1)} - \mathcal{Y}^{(1)} \|_{F}^{2} 
  + \frac{\beta}{2}\sum_{k=2}^{3}\| \mathcal{T}^{(2)} - \mathcal{Y}^{(k)} \|_{F}^{2} \label{eq:coup_objective_overlap_legrange_1}
\end{multline} 
where $\beta$ is a proximity parameter. Using this Lagrangian formulation, we can obtain solutions for unknown variables $M$, $\mathcal{T}^{(1)}$, $\mathcal{T}^{(2)}$, $W^{M}$, $\mathcal{W}^{\mathcal{T}^{(k)}} \; (k = 1,2,3)$, $X$, and $\mathcal{Y}^{(k)} \; (k = 1,2,3)$ iteratively. We use superscripts $[t]$ and $[t-1]$ to represent the variables at iteration steps $t$ and $t-1$, respectively.     

The solutions for $M$ at each iteration can be obtained by solving the following sub-problem:
\begin{equation*}
 M^{[t]}  = unvec \bigg( ( \Omega_{M}^{\top}\Omega_{M} + \beta I_{M})^{-1}vec\big(\Omega_{M}(\hat{M}) - W^{M[t-1]} + \beta X^{[t-1]} \big) \bigg). 
\end{equation*}

Solutions for $\mathcal{T}^{(1)}$ and  $\mathcal{T}^{(2)}$ at  iteration step $t$ can be obtained from the following sub-problem:
\begin{multline}
\begin{bmatrix} 
\Omega_{\mathcal{T}}^{\top}\Omega_{\mathcal{T}} + 2\beta I_{\mathcal{T}} & I_{\mathcal{T}} \\ I_{\mathcal{T}} & \Omega_{\mathcal{T}}^{\top}\Omega_{\mathcal{T}} + 2\beta I_{\mathcal{T}} 
\end{bmatrix}
\left[ \begin{array}{c} vec(\mathcal{T}^{(1)[t]}) \\ vec(\mathcal{T}^{(2)[t]}) \end{array} \right] =\\
\\ \left[ \begin{array}{c} vec \bigg(\Omega_{\hat{\mathcal{T}}} (\hat{\mathcal{T}}) - \sum_{k=2}^{3} \mathcal{W}^{\mathcal{T}^{(k)}[t-1] } + \beta \sum_{k=2}^{3} \mathcal{Y}^{(k)[t-1]}  \bigg) \\ vec \bigg(\Omega_{\hat{\mathcal{T}}} (\hat{\mathcal{T}}) - \sum_{k=2}^{3} \mathcal{W}^{\mathcal{T}^{(k)}[t-1]} + \beta \sum_{k=2}^{3} \mathcal{Y}^{(k)[t-1]}  \bigg) \end{array} \right] ,\label{eq:coup_tensor_subproblem}
\end{multline} 
where $I_{M}$ and $I_{\mathcal{T}}$ are unit diagonal matrices with dimensions  $n_{1}m \times n_{1}m$ and $n_{1}n_{2}n_{3} \times n_{1}n_{2}n_{3}$, respectively.

The updates for $X$ and $\mathcal{Y}^{(k)}$, $(k = 1,2,3)$ at  iteration step $t$ are given as
\begin{equation}
[Y_{(1)}^{(1)[t-1]};X^{[t-1]}] = {\rm{prox}}_{{\lambda/(\sqrt{n_{1}}\beta)} }\bigg([\frac{W_{(1)}^{\mathcal{T}^{(1)}[t-1]}}{\beta} ; \frac{W^{M[t-1]}}{\beta} ] +  [ T_{(1)}^{(1)[t]} ; M^{[t]} ]  \bigg), \label{eq:coup_tensor_subproblem_prox1}
\end{equation}
and
\begin{equation}
Y_{(k)}^{(k)[t-1]} = {\rm{prox}}_{{\lambda/\beta} }\bigg(\frac{W_{(k)}^{\mathcal{T}^{(t)}[t-1]}}{\beta} +  T_{(k)}^{(2)[t]} \bigg), \quad k = 2,3, \label{eq:coup_tensor_subproblem_prox2}
\end{equation}
where $\rm{prox}_{\lambda}(X) = U(S-\lambda)_{+}V^{\top}$ for $X = USV^{\top}$.

The update rules for the dual variables are
\begin{equation*}
W^{M[t]} = W^{M[t-1]} + \beta(M^{[t]} -X^{[t]}), 
\end{equation*}
\begin{equation*}
\mathcal{W}^{\mathcal{T}^{(1)}[t-1]} = \mathcal{W}^{\mathcal{T}^{(1)}[t]} + \beta(\mathcal{T}^{(1)[t]} - \mathcal{Y}^{(1)[t]}),  
\end{equation*}  
\begin{equation*}
\mathcal{W}^{\mathcal{T}^{(k)}[t-1]} = \mathcal{W}^{\mathcal{T}^{(k)}[t]} + \beta(\mathcal{T}^{(k)[t]} - \mathcal{Y}^{(k)[t]}), \quad k =2,3.  
\end{equation*} 

We can modify the above optimization procedures by replacing the variables in \eqref{eq:coup_objective_overlap} in accordance with the norm that is used to  regularize the tensor and by adjusting operations in \eqref{eq:coup_objective_overlap_legrange},  \eqref{eq:coup_tensor_subproblem}, \eqref{eq:coup_tensor_subproblem_prox1}, and \eqref{eq:coup_tensor_subproblem_prox2}. For  example, for the norm $\| \cdot \|^{1}_{(\mathrm{O},\mathrm{O},\mathrm{O})}$, there is only a single $\mathcal{T}$, so the  sub-problem for \eqref{eq:coup_tensor_subproblem} becomes
\begin{equation*}
\big( \Omega_{\mathcal{T}}^{\top}\Omega_{\mathcal{T}} + 3\beta I_{\mathcal{T}}  \big)
 vec(\mathcal{T}^{[t]}) 
=
  vec \bigg( \Omega_{\hat{\mathcal{T}}} (\hat{\mathcal{T}}) - \sum_{k=1}^{3} \mathcal{W}^{\mathcal{T}^{(k)}[t-1]}  + \beta \sum_{k=1}^{3} \mathcal{Y}^{[t-1]}  \bigg),
\end{equation*}
and that for \eqref{eq:coup_tensor_subproblem_prox1} becomes
\begin{equation*}
[Y_{(1)}^{(1)[t]};X^{[t]}] = {\rm{prox}}_{{\lambda/\beta} }\bigg([\frac{W_{(1)}^{\mathcal{T}^{(k)}[t-1]}}{\beta} ; \frac{W^{M[t-1]}}{\beta} ] +  [ T_{(1)}^{[t]} ; M^{[t]} ]  \bigg),
\end{equation*}
and
\begin{equation*}
Y_{(k)}^{(k)[t-1]} = {\rm{prox}}_{{\lambda/\beta} }\bigg(\frac{W_{(k)}^{\mathcal{T}^{(k)}[t-1]}}{\beta} +  T_{(k)}^{[t]} \bigg), \quad k = 1,2,3.
\end{equation*}
Additionally, the dual update rule with $\mathcal{T}$ becomes
\begin{equation*}
\mathcal{W}^{\mathcal{T}^{(k)}[t-1]} = \mathcal{W}^{\mathcal{T}^{(k)}[t]} + \beta(\mathcal{T}^{[t]} - \mathcal{Y}^{(k)[t]}), \quad k =1,2,3.  
\end{equation*} 
The optimization procedures for the other norms can be similarly derived.

\section{Theoretical Analysis}
In this section, we analyze the excess risk bounds of the completion model  introduced in \eqref{eq:coup_objective} for the coupled norms defined in Section 3 using  transductive Rademacher complexity \citep{transductive_rademacher,JMLR:v15:shamir14a}. Let us again consider matrix $M$ and tensor $\mathcal{T}$  and use them as a single structure $\bold{X} = \mathcal{T}\cup M$  with a training sample index set  $\mathrm{S}_{\mathrm{Train}}$ and a testing sample index set  $\mathrm{S}_{\mathrm{Test}}$ with the total set of observed samples $\mathrm{S} = \mathrm{S}_{\mathrm{Train}}\cup\mathrm{S}_{\mathrm{Test}}$.  We rewrite   \eqref{eq:coup_objective} with our new notations as an equivalent model:
\begin{multline}
\underset{\bold{W}}{\min} \frac{1}{|\mathrm{S}_{\mathrm{Train}} |} \sum_{(i_{1},i_{2},i_{3}) \in \mathrm{S}_{\mathrm{Train}} } l(\bold{X}_{i_{1},i_{2},i_{3}}, \bold{W}_{i_{1},i_{2},i_{3}}) 
\;\; \mathrm{s.t.} \;\;  \| \bold{W} \|_{\mathrm{cn}} \leq B \label{eq:coup_objective_thory}, 
\end{multline}
where $l(a,b) = (a-b)^{2}$, $\bold{W} = \mathcal{W}\cup W_{M}$ is the learned coupled structure consisting of components $\mathcal{W}$ and $W_{M}$ of the tensor and matrix, respectively, $B$ is a constant, and $\| \cdot \|_{\mathrm{cn}}$ is  any norm defined in  Section 3.2.  

Given that $l(\cdot,\cdot)$  is a $\Lambda$-Lipschitz  loss function  bounded by $\sup_{i_{1},i_{2},i_{3} } |l(\bold{X}_{i_{1},i_{2},i_{3}}, \bold{W}_{i_{1},i_{2},i_{3}})| \leq b_{l}$ and assuming that  $|\mathrm{S}_{\mathrm{Train}}| = |\mathrm{S}_{\mathrm{Test}}| = |S|/2$, we can obtain the following excess risk bound based on transductive Rademacher complexity theory \citep{transductive_rademacher,JMLR:v15:shamir14a} with probability $1-\delta$, 
\begin{multline}
\frac{1}{|\mathrm{S}_{\mathrm{Test}}|} \sum_{(i_{1},i_{2},i_{3}) \in \mathrm{S}_{\mathrm{Test}}} l(\bold{X}_{i_{1},i_{2},i_{3}}, \bold{W}_{i_{1},i_{2},i_{3}}) \\ 
-  \frac{1}{|\mathrm{S}_{\mathrm{Train}}|} \sum_{(i_{1},i_{2},i_{3}) \in \mathrm{S}_{\mathrm{Train}}} l(\bold{X}_{i_{1},i_{2},i_{3}}, \bold{W}_{i_{1},i_{2},i_{3}})  \\ 
\leq  4R(\bold{W}) + b_{l}\bigg(\frac{11 + 4\sqrt{\log{\frac{1}{\delta}}}}{\sqrt{|\mathrm{S}_{\mathrm{Train}} |}}\bigg),\label{eq:transductive_formulation}
\end{multline} 
where  $R(\bold{W})$ is the transductive Rademacher complexity  defined as
\begin{equation}
R(\bold{W}) = 
 \frac{1}{|\mathrm{S}|}\mathbb{E}_{\sigma} \bigg[\sup_{\|\bold{W}\|_{\mathrm{cn}} \leq B}  \sum_{(i_{1},i_{2},i_{3}) \in \mathrm{S}} \sigma_{i_{1},i_{2},i_{3} }l(\bold{W}_{i_{1},i_{2},i_{3} } ,\bold{X}_{i_{1},i_{2},i_{3} } ) \bigg], \label{eq:rademacher_complexity}
\end{equation}
where $\sigma_{i_{1},i_{2},i_{3}} \in \{-1,1\} $  with probability $0.5$ if $(i_{1},i_{2},i_{3}) \in \mathrm{S}$, or $0$ otherwise (See Appendix B for derivation).

Next we give the bounds for \eqref{eq:rademacher_complexity} with respect to different coupled norms. We assume that $|\mathrm{S}_{\mathrm{Train}}| = |\mathrm{S}_{\mathrm{Test}}|$, as in \citep{JMLR:v15:shamir14a}, but our theorem can be extended to more general cases. Detailed proofs of the theorems in this section are given in Appendix B.  
 
The following two theorems give the Rademacher complexities for coupled completion regularized using the coupled norms  $\|\cdot \|^{1}_{(\mathrm{O},\mathrm{O},\mathrm{O})}$   and $\| \cdot \|^{1}_{(\mathrm{S},\mathrm{S},\mathrm{S})}$.   
\begin{thm} 
Let $\| \cdot \|_{\mathrm{cn}} = \| \cdot \|^{1}_{(\mathrm{O},\mathrm{O},\mathrm{O})}$; then, with  probability $1-\delta$,
\begin{equation*}
\begin{split}
R(\bold{W}) &\leq \frac{3\Lambda}{2|\mathrm{S}|} \Big[  \sqrt{r_{(1)}}(B_{\mathcal{T}} + B_{M}) + \sum_{k=2}^{3}\sqrt{r_{k}}B_{\mathcal{T}} \Big]
\\ 
& \qquad\qquad \max \Bigg\{  C_{2}\bigg(\sqrt{n_{1}} + \sqrt{\prod_{j=2}^{3}{n_{j}} + m} \bigg), \min_{k \in {2,3}} C_{1} \bigg( \sqrt{n_{k}} +  \sqrt{\prod_{j \neq k}^{3}{n_{j}}} \bigg)  \Bigg\},
\end{split}
\end{equation*}
where  $(r_{1},r_{2},r_{3})$ is the multilinear rank of $\mathcal{W}$, $r_{(1)}$ is the rank of the coupled unfolding on mode $1$, and $B_{M}$, $B_{T}$, $C_{1}$, and $C_{2}$ are constants.
\end{thm}

\begin{thm} 
Let $\| \cdot \|_{\mathrm{cn}} = \| \cdot \|^{1}_{(\mathrm{S},\mathrm{S},\mathrm{S})}$,   then, with  probability $1-\delta$,
\begin{equation*}
\begin{split}
R(\bold{W}) &\leq \frac{3\Lambda}{2|\mathrm{S}|} \Bigg[ \sqrt{\frac{r_{(1)}}{n_{1}}}(B_{M} + B_{\mathcal{T}}) + \min_{k \in 2,3} \sqrt{\frac{r_{k}}{n_{k}}}B_{\mathcal{T}} \Bigg]\\
 & \qquad \max \Bigg\{ C_{2}\Bigg(n_{1} + \sqrt{\prod_{i=1}^{3}{n_{i}} + n_{1}m }\Bigg) , C_{1} \max_{k=2,3}\Bigg(n_{k} + \sqrt{\prod_{i = 1}^{3}{n_{i}}}\Bigg) \Bigg\},
\end{split}
\end{equation*}
where  $(r_{1},r_{2},r_{3})$ is the multilinear rank of $\mathcal{W}$, $r_{(1)}$ is the rank of the coupled unfolding on mode $1$, and $B_{M}$, $B_{T}$, $C_{1}$, and $C_{2}$ are constants.
\end{thm} 

We can see that in both of these theorems, the Rademacher complexity of the coupled tensor is divided by the total number of observed samples of both the matrix and the tensor. If the tensor or the matrix is completed separately, then the Rademacher complexity is only divided by their individual samples (see Theorems 7--9 in  the Appendix B and a discussion elsewhere \citep{JMLR:v15:shamir14a}). This means that coupled tensor learning can lead  to better performance than separate  matrix or  tensor learning. We can also see that, due to coupling, the excess risks are bounded by the ranks of both the tensors and the concatenated matrix of  the unfolded tensors on  the coupled mode. Additionally, the maximum term on the right takes the combinations of  both the tensor and the concatenated matrix of the unfolded tensors on the coupled mode. 

Finally, we consider the Rademacher complexity of the mixed norm $\| \cdot \|_{\mathrm{cn}} = \| \cdot \|^{1}_{(\mathrm{S},\mathrm{O},\mathrm{O})}$.
\begin{thm} 
Let $\| \cdot \|_{\mathrm{cn}} = \| \cdot \|^{1}_{(\mathrm{S},\mathrm{O},\mathrm{O})}$; then, with  probability $1-\delta$,
\begin{equation*}
\begin{split}
R(\bold{W}) &\leq \frac{3\Lambda}{2|\mathrm{S}|} \Bigg[\sqrt{\frac{r_{(1)}}{n_{1}}}(B_{M} + B_{\mathcal{T}})  +  \sum_{i=2,3} \sqrt{r_{i}}  B_{\mathcal{T}}\Bigg] \\
 & \qquad \max  \Bigg\{ C_{2}\Bigg( n_{1} + \sqrt{\prod_{i=1}^{3}n_{i} + n_{1}m} \Bigg),  \min_{k=2,3} C_{1} \Bigg(\sqrt{n_{k}} + \sqrt{\prod_{i \neq k}^{3}n_{i}}\Bigg) \Bigg\},
\end{split}
\end{equation*}
where  $(r_{1},r_{2},r_{3})$ is the multilinear rank of $\mathcal{W}$, $r_{(1)}$ is the rank of the coupled unfolding on mode $1$, and $B_{M}$, $B_{\mathcal{T}}$, $C_{1}$, and $C_{2}$ are constants.
\end{thm} 

We see that, for the mixed norm $\| \cdot \|_{\mathrm{cn}} = \| \cdot \|^{1}_{(\mathrm{S},\mathrm{O},\mathrm{O})}$, the excess risk is bounded by the scaled rank of the coupled unfolding along the first mode. For this norm, we can see that the terms related to ranks are smaller in Theorem 3 and that the maximum term could be smaller than in Theorem 4. This means that this norm can perform better than $\| \cdot \|^{1}_{(\mathrm{O},\mathrm{O},\mathrm{O})}$ and $\| \cdot \|^{1}_{(\mathrm{S},\mathrm{S},\mathrm{S})}$ depending on the ranks and mode dimensions of the coupled tensor. The bounds of the other two mixed norms can also be derived and explained in a manner similar to Theorem 5.

\section{Evaluation}
We evaluated our proposed method experimentally using synthetic and real-world data. 

\subsection{Synthetic Data}
Our main objectives  were to evaluate how the  proposed norms perform depending on the ranks and dimensions of the coupled tensors. We used simulation data based on CP rank and Tucker rank in these experiments.

\subsubsection{Experiments Using CP Rank}
To create coupled tensors with the CP rank, we first generated a $3$-mode tensor $\mathcal{T} \in \mathbb{R}^{n_{1} \times n_{2} \times n_{3} }$ with CP rank  $r$ using CP decomposition  \citep{journals/siamrev/KoldaB09} as $\mathcal{T} = \sum_{i=1}^{r} c_{i} u_{i}  \circ v_{i} \circ w_{i}$ where $u_{i} \in \mathbb{R}^{n_{1}}$, $v_{i} \in \mathbb{R}^{n_{2}}$ and $w_{i} \in \mathbb{R}^{n_{3}}$  and $c_{i} \in \mathbb{R}^{+} $. For our experiments, we used two approaches to create CP-rank-based tensors in which all the component vectors $u_{i}, v_{i}$, and $w_{i}$ were nonorthogonal vectors or orthogonal vectors.   We coupled  matrix $X \in \mathbb{R}^{n_{1} \times m}$ with rank $r$ to $\mathcal{T}$ on  mode $1$ by generating $X = USV^{\top}$ with $U(1:r,:) =[u_{1},\ldots,u_{r}]$, $S \in \mathbb{R}^{r \times r}$, and $V \in \mathbb{R}^{m \times r}$ is an orthogonal matrix.   We also added  noise  sampled from a Gaussian distribution with mean zero and variance of $0.01$ to the elements of the matrix and the tensor.

In our experiments using synthetic data, we considered  coupled structures of tensors with dimension  $20 \times 20 \times 20$ and  matrices with dimension  $20 \times 30$ coupled on their first modes.  To simulate completion, we randomly selected observed samples with percentages of  $30$, $50$, and $70$ of the total number of elements in both the matrix and the tensor, selected a validation set with a percentage of $10$ and took the remainder as test samples. We performed coupled completion using the  proposed coupled norms of $\| \cdot \|_{(\mathrm{O},\mathrm{O},\mathrm{O})}^{1}$, $\| \cdot \|_{(\mathrm{S},\mathrm{S},\mathrm{S})}^{1}$, $\| \cdot \|_{(\mathrm{S},\mathrm{O},\mathrm{O})}^{1}$, $\| \cdot \|_{(\mathrm{O},\mathrm{S},\mathrm{O})}^{1}$, and $\| \cdot \|_{(\mathrm{O},\mathrm{O},\mathrm{S})}^{1}$. For all the learning models with these norms, we cross-validated their regularization parameters ranging from $0.01$ to $5.0$ with intervals of $0.05$. We ran our experiments with $10$ random selections and plotted the mean square error (MSE) for the test samples.

As benchmark methods  we used the overlapped trace norm (OTN) and the scaled latent trace norm (SLTN) for individual tensors and the matrix trace norm (MTN) for individual matrices. For all these norms, we cross-validated  the regularization parameters ranging from $0.01$ to $5.0$ with intervals of $0.05$. We  compared our results with those of  advanced coupled matrix-tensor factorization  ACMTF  \citep{DBLP:journals/bmcbi/AcarPGRLNB14}, for which the regularization parameters were selected using cross-validation in the range $0, 0.0001, 0.001, \ldots,1$. To select ranks to use with the ACMTF method, we first ran experiments using ranks of $1,3,5,\dots,19$ and selected the rank that gave the best performance. Due to the non-convex nature of ACMTF, we ran experiments with $5$ random initializations to select the best local optimal solution.

\begin{figure*}[h]
 \centering
 \begin{minipage}[t]{0.49\linewidth}
\centering
  {\includegraphics[width=0.99\textwidth]{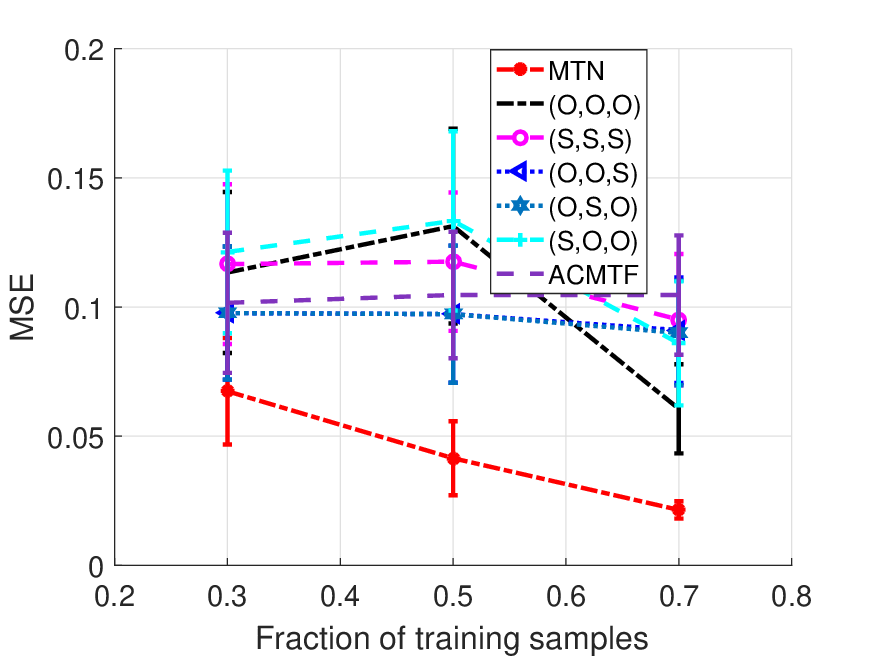}} \\ \vspace{-0.10cm}
(a) Matrix Completion 
\end{minipage}
 \begin{minipage}[t]{0.49\linewidth}
\centering
  {\includegraphics[width=0.99\textwidth]{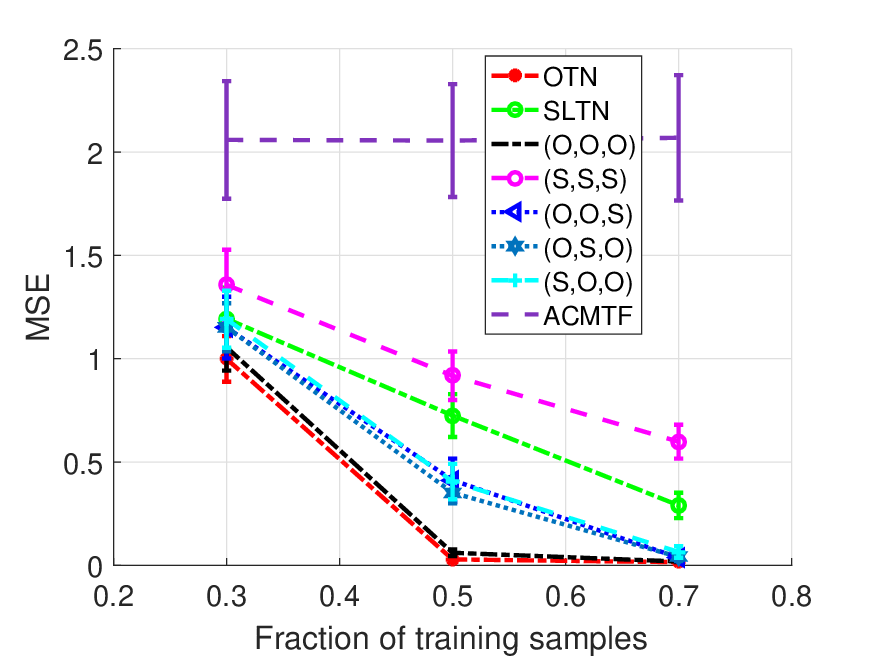}} \\ \vspace{-0.10cm}
(b) Tensor Completion. 
\end{minipage}
\caption{Completion performance of matrix with dimension $20 \times 30$ and rank  $5$ with no sharing and of  tensor with dimension  $20 \times 20 \times 20$ and CP rank  $5$ with nonorthogonal component vectors.}
  \label{fig:cp-5-5-nonorth}
\end{figure*}

We first ran experiments on coupled tensor completion based on CP rank in different settings. In the first experiment, we considered coupled tensors with no shared components.   
In this experiment, we created a tensor with CP rank $5$ in which the component vectors  were nonorthogonal and generated from a normal distribution. We also created a matrix of rank $5$ and without any components in common with the tensor. Figure \ref{fig:cp-5-5-nonorth} shows that the coupled norms did not perform better than individual matrix completion using the matrix trace norm. However, for tensor completion, the coupled norm $\| \cdot \|_{(\mathrm{O},\mathrm{O},\mathrm{O})}^{1}$ had performance comparable to that of the overlapped trace norm.     

\begin{figure*}[h]
 \centering
 \begin{minipage}[t]{0.49\linewidth}
\centering
  {\includegraphics[width=0.99\textwidth]{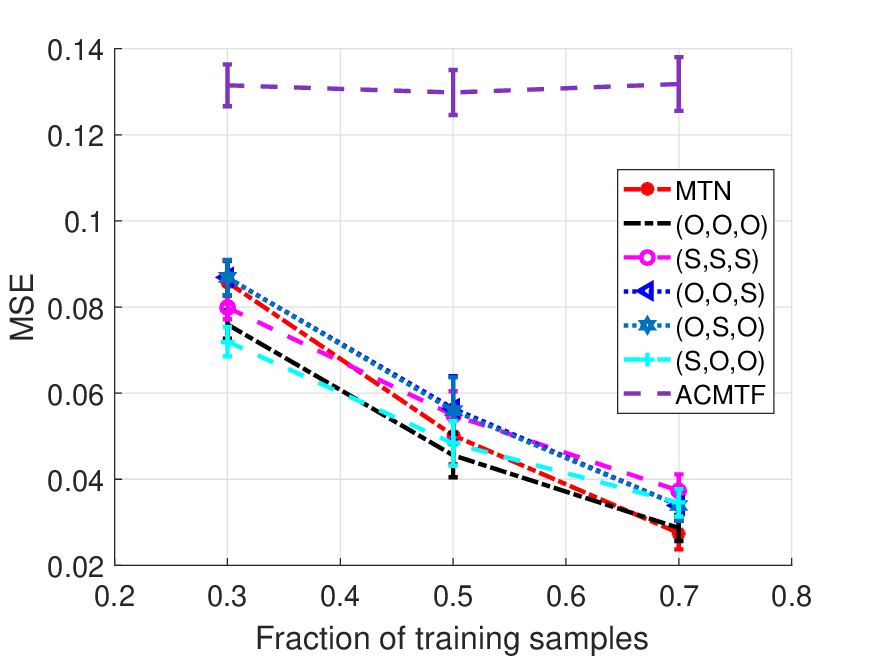}} \\ \vspace{-0.10cm}
(a) Matrix Completion 
\end{minipage}
 \begin{minipage}[t]{0.49\linewidth}
\centering
  {\includegraphics[width=0.99\textwidth]{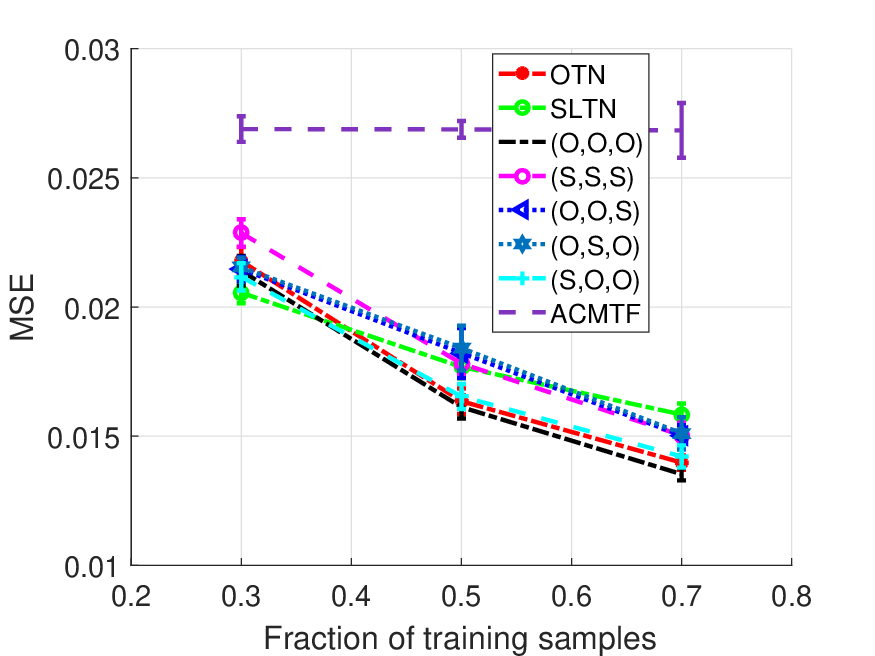}} \\ \vspace{-0.10cm}
(b) Tensor Completion. 
\end{minipage}
\caption{Completion performance of matrix with dimension $20 \times 30$ and rank $5$ with all components shared and of tensor with dimension $20 \times 20 \times 20$ and CP rank  $5$ with orthogonal component vectors.}
  \label{fig:cp-5-5}
\end{figure*}

We next ran experiments on coupled tensors with some components in common and with both orthogonal and nonorthogonal component vectors. We created coupled tensors with CP rank of $5$, and both  the tensor and matrix shared all components along mode $1$. We  generated the tensor with orthogonal component vectors. As shown in Figure \ref{fig:cp-5-5}, the coupled norm $\| \cdot \|_{(\mathrm{O},\mathrm{O},\mathrm{O})}^{1}$ had good performance for both the matrix and tensor.

\begin{figure*}[h]
 \centering
 \begin{minipage}[t]{0.49\linewidth}
\centering
  {\includegraphics[width=0.99\textwidth]{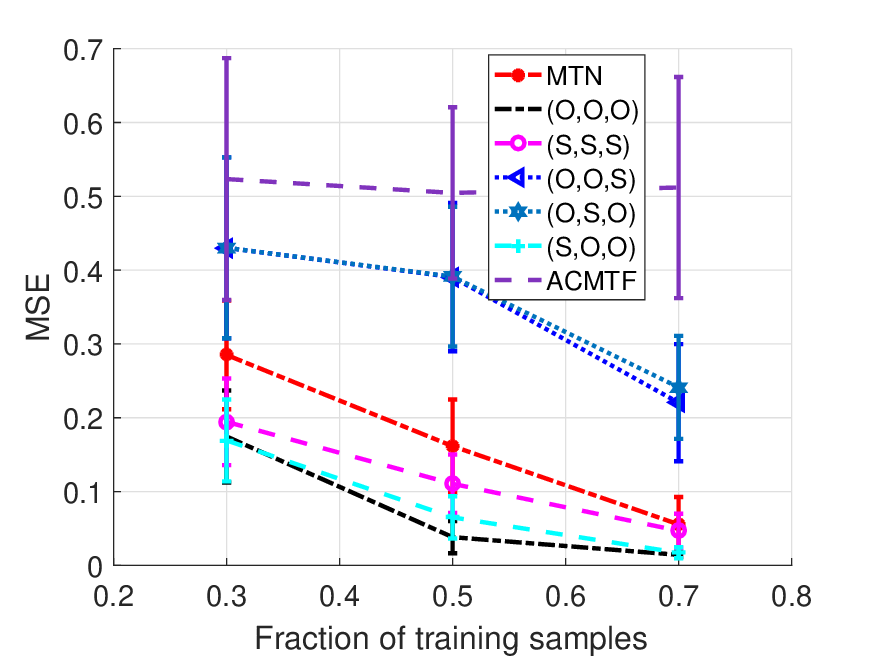}} \\ \vspace{-0.10cm}
(a) Matrix Completion 
\end{minipage}
 \begin{minipage}[t]{0.49\linewidth}
\centering
  {\includegraphics[width=0.99\textwidth]{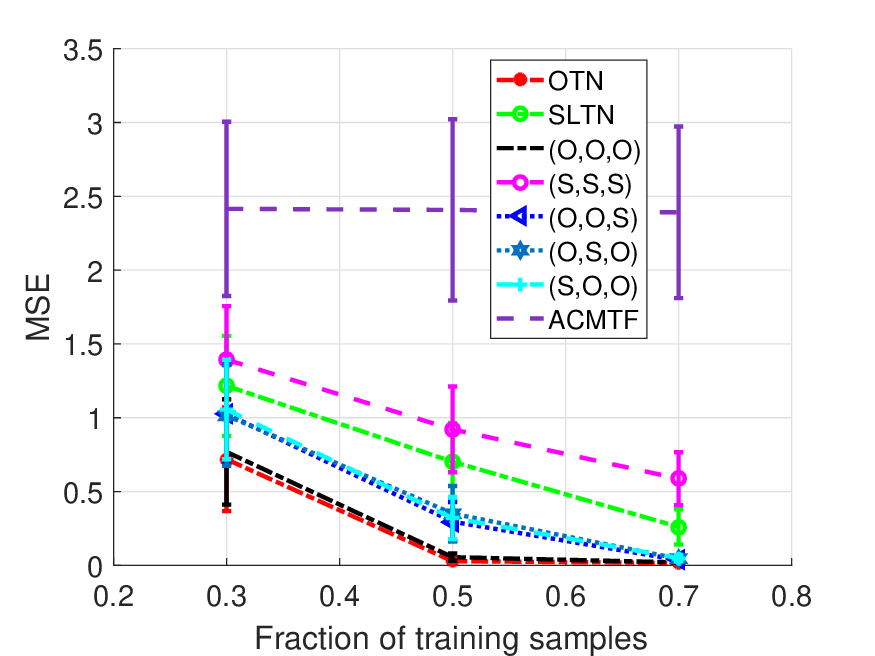}} \\ \vspace{-0.10cm}
(b) Tensor Completion. 
\end{minipage}
\caption{Completion performance of matrix with dimension $20 \times 30$ and rank of $5$ with all  component vectors shared and of tensor with dimension $20 \times 20 \times 20$ and CP rank  $5$ and nonorthogonal component vectors.}
  \label{fig:cp-5-5_nonorth}
\end{figure*}

Figure \ref{fig:cp-5-5_nonorth}, we shows the performance of coupled tensors with the same rank as in the previous experiment with  tensors created from nonorthogonal component vectors.  Again the coupled norm $\| \cdot \|_{(\mathrm{O},\mathrm{O},\mathrm{O})}^{1}$ had  better performance than individual matrix and tensor completions.  

\begin{figure*}[t]
 \centering
 \begin{minipage}[t]{0.49\linewidth}
\centering
  {\includegraphics[width=0.99\textwidth]{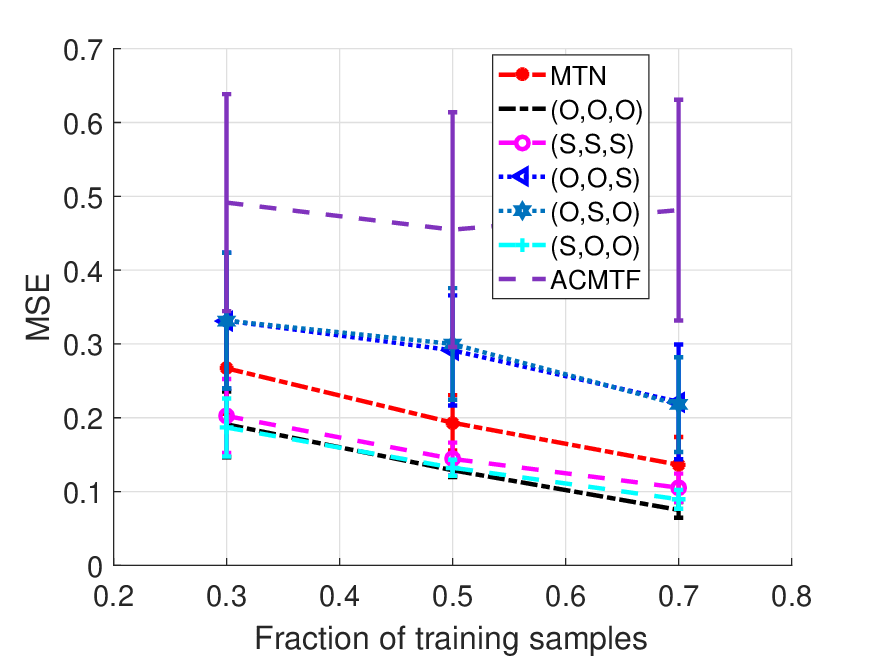}} \\ \vspace{-0.10cm}
(a) Matrix Completion 
\end{minipage}
 \begin{minipage}[t]{0.49\linewidth}
\centering
  {\includegraphics[width=0.99\textwidth]{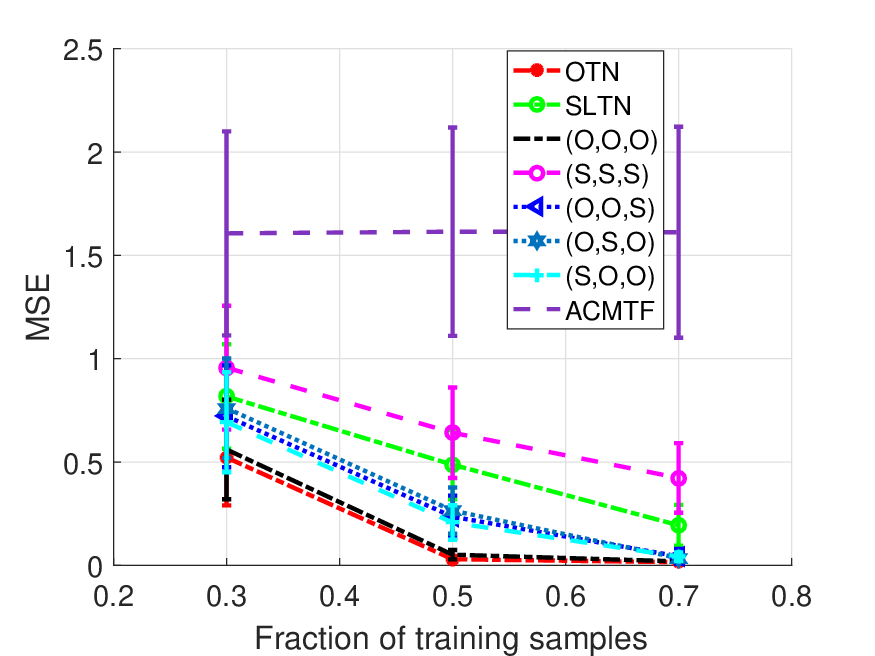}} \\ \vspace{-0.10cm}
(b) Tensor Completion. 
\end{minipage}
\caption{Completion performance of matrix with dimension $20 \times 30$ and rank $5$ and of tensor with dimension  $20 \times 20 \times 20$ with CP rank $10$ and nonorthogonal component vectors that shared $5$ components.}
  \label{fig:cp-5-10_nonorth}
\end{figure*}

\begin{figure*}[t]
 \centering
 \begin{minipage}[t]{0.49\linewidth}
\centering
  {\includegraphics[width=0.99\textwidth]{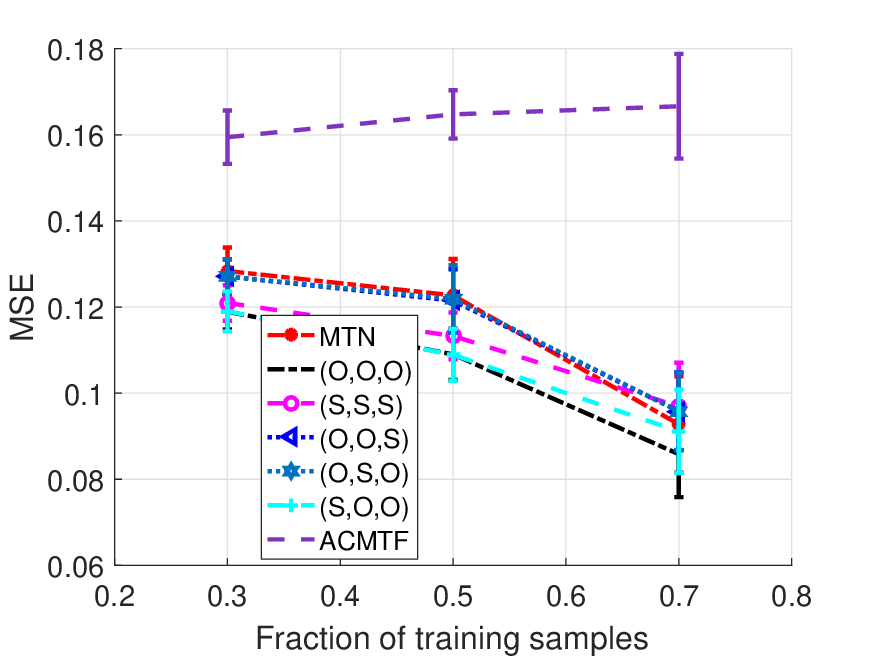}} \\ \vspace{-0.10cm}
(a) Matrix Completion 
\end{minipage}
 \begin{minipage}[t]{0.49\linewidth}
\centering
  {\includegraphics[width=0.99\textwidth]{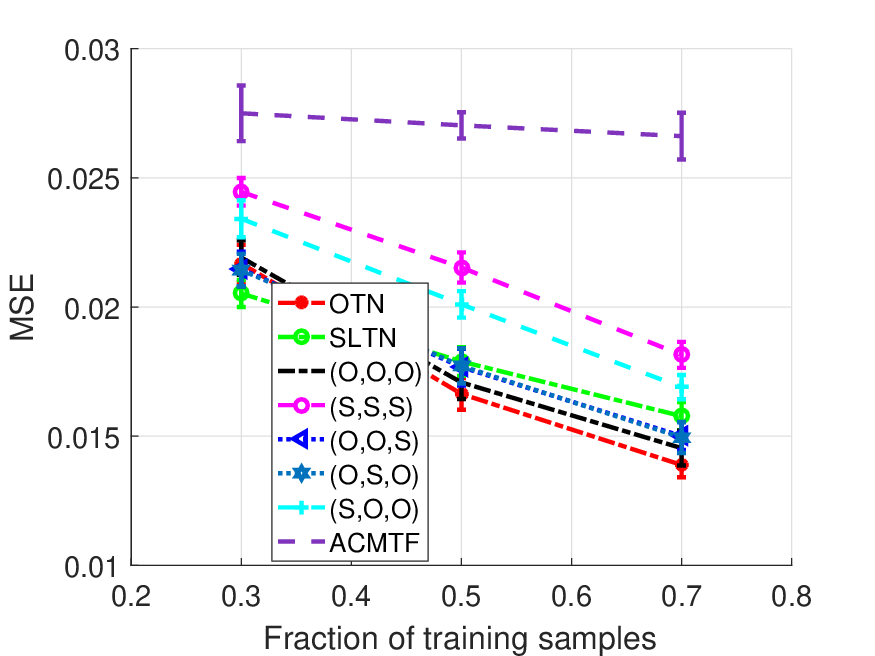}} \\ \vspace{-0.10cm}
(b) Tensor Completion. 
\end{minipage}
\caption{Completion performance of matrix with dimension $20 \times 30$ and rank $5$ and of tensor with dimension $20 \times 20 \times 20$ and CP rank $10$ and orthogonal component vectors that shared $5$ components.}
  \label{fig:cp-5-10}
\end{figure*}

In our final experiment, we created tensors with CP rank $5$  and coupled them with a matrix of rank $10$ sharing all $5$ component vectors along mode $1$. Figures \ref{fig:cp-5-10_nonorth} and \ref{fig:cp-5-10} show the results for tensors created with orthogonal and nonorthogonal component vectors, respectively. In both cases, the coupled norms $\| \cdot \|_{(\mathrm{O},\mathrm{O},\mathrm{O})}^{1}$ ,  $\| \cdot \|_{(\mathrm{S},\mathrm{S},\mathrm{S})}^{1}$, and $\| \cdot \|_{(\mathrm{S},\mathrm{O},\mathrm{O})}^{1}$ had better matrix completion performance than individual completion by the matrix trace norm. Similarly, as in the previous experiments, both the overlapped trace norm and the coupled norm $\| \cdot \|_{(\mathrm{O},\mathrm{O},\mathrm{O})}^{1}$ had comparable performances.

\subsubsection{Simulations Using Tucker Rank}
To create coupled tensors with the Tucker rank, we first generated a tensor $\mathcal{T} \in \mathbb{R}^{n_{1} \times n_{2} \times n_{3} }$ using  Tucker decomposition \citep{journals/siamrev/KoldaB09} as $\mathcal{T} = \mathcal{C} \times_{1} U_{1} \times_{2} U_{2} \times_{3} U_{3}$, where $\mathcal{C} \in  \mathbb{R}^{r_{1} \times r_{2} \times r_{3} }$ was the core tensor generated from a normal distribution specifying multilinear rank $(r_{1},r_{2},r_{3})$ and  component matrices  $U_{1} \in \mathbb{R}^{r_{1} \times p_{1}}$, $U_{2} \in \mathbb{R}^{r_{2} \times p_{2}}$, and $U_{3} \in \mathbb{R}^{r_{3} \times p_{3}}$ were orthogonal matrices. Next we generated a matrix that was  coupled with mode $1$ of the tensor  using singular value decomposition $X = USV^{\top}$, where we specified its rank $r$  using diagonal matrix $S$ and generated matrices $U$ and $V$ as orthogonal matrices. For sharing between the matrix and the tensor, we computed $T_{(1)} = U_{n}S_{n}V^{\top}_{n}$,  and  replaced the first $s$ singular values of $S$ with the first $s$ singular values of $S_{n}$, replaced the first basis vectors  $s$ of $U$ with the first  $s$ basis vectors of $U_{n}$, and computed $X = USV^{\top}$ such that the coupled structure shared $s$ common components. We also added  noise  sampled from a Gaussian distribution with mean  zero and variance  $0.01$ to  the elements of the coupled tensor. 

As in the synthetic experiments using the CP rank, we considered  coupled structures with tensors with dimension  $20 \times 20 \times 20$ and  matrices with dimension  $20 \times 30$ coupled on their mode $1$. We considered different multilinear ranks of tensors, ranks of matrices, and  degrees of sharing among them. We used the same percentages in selecting the training, testing, and validation sets as we did in the  CP rank  experiments. We  again compared our results with those of ACMTF. 

We also used an additional non-convex coupled learning model to incorporate multilinear ranks of the coupled tensor by considering  Tucker decomposition under the assumption that the components of the coupled mode were shared between both the matrix and tensor. We used the Tensorlab framework \citep{tensorlab3.0} to implement this model. We  regularized the factorized components of the tensor (including the core tensor) and the matrix using the Frobenius norm. We used a regularization parameter selected from the range $0.01$ to $50$ in logarithmic linear scale with $5$ divisions (in Matlab syntax \texttt{exp(linspace(log(0.01), log(50), 5))}). We refer to this benchmark method as NC-Tucker. Due to the non-convex nature of the model, we ran   $5-10$ simulations with different random initializations and selected the best local optimal solution. Specifying the multilinear rank a priori for this model would be challenging in real applications, but since we knew the rank in our simulations,  we could specify the multilinear ranks to be used to create the tensors.

\begin{figure*}[h]
 \centering
 \begin{minipage}[t]{0.46\linewidth}
\centering
  {\includegraphics[width=0.99\textwidth]{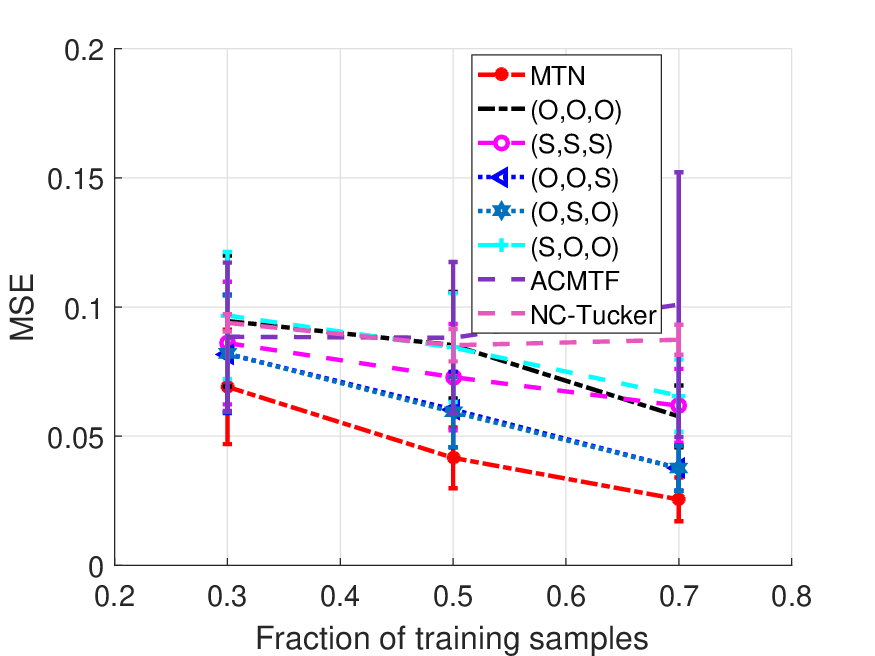}} \\ \vspace{-0.10cm}
(a) Matrix Completion 
\end{minipage}
 \begin{minipage}[t]{0.46\linewidth}
\centering
  {\includegraphics[width=0.99\textwidth]{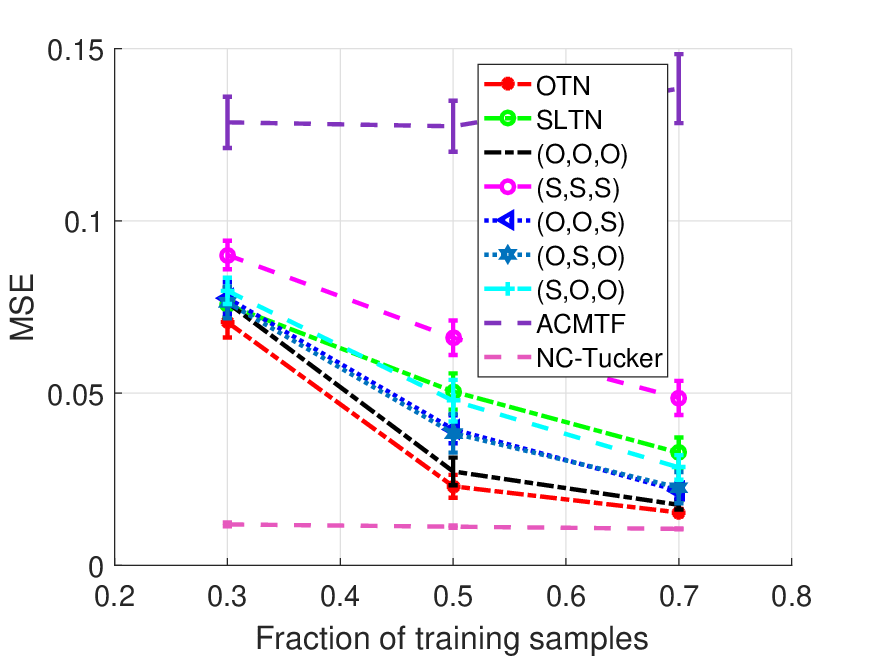}} \\ \vspace{-0.10cm}
(b) Tensor Completion. 
\end{minipage}
\caption{Completion performance of matrix with dimension  $20 \times 30$ and rank $5$ and of tensor with dimension $20 \times 20 \times 20$ and multilinear rank $(5, 5, 5)$  with no sharing.}
  \label{fig:202020-2030-5-555}
\end{figure*}

In our first simulations,  we considered a coupled tensor  with a matrix  rank of $5$ and a tensor  multilinear rank  $(5,5,5)$ with no shared components.  Figure \ref{fig:202020-2030-5-555} shows that, with this setting, individual matrix and tensor completion had better performance than that of the coupled norms. The non-convex NC-Tucker benchmark method had  the best performance for the tensor but performed poorly in matrix completion compared to the coupled norms.

\begin{figure*}[h!]
 \centering
  \begin{minipage}[t]{0.46\linewidth}
\centering
  {\includegraphics[width=0.99\textwidth]{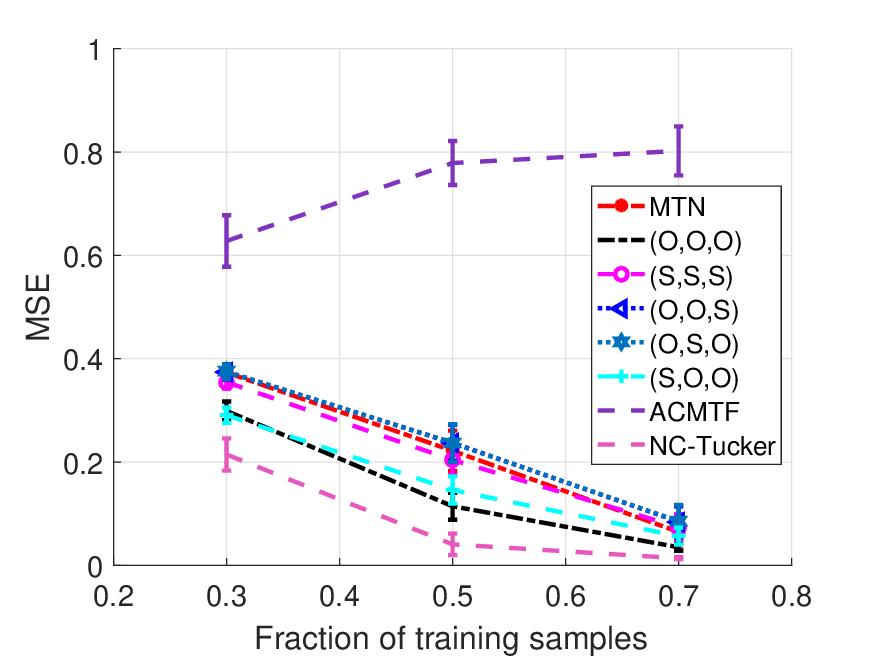}} \\ \vspace{-0.10cm}
(a) Matrix Completion 
\end{minipage}
 \begin{minipage}[t]{0.46\linewidth}
\centering
  {\includegraphics[width=0.99\textwidth]{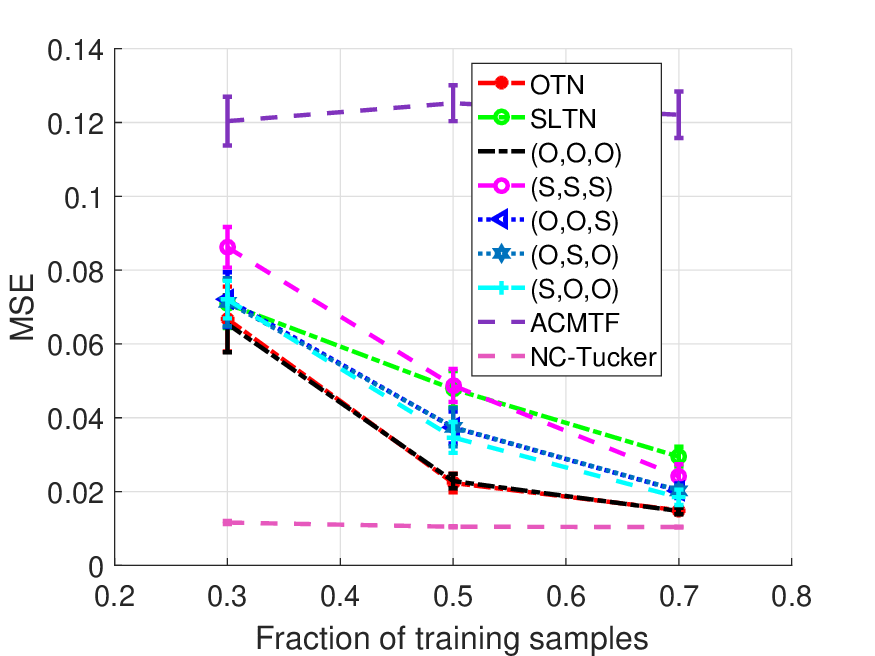}} \\ \vspace{-0.10cm}
(b) Tensor Completion.
\end{minipage}
\caption{Completion performances of completion of matrix with dimension  $20 \times 30$ and rank $5$ and of tensor with dimension $20 \times 20 \times 20$ and multilinear rank $(5, 5, 5)$ that shared $5$ components.}
\label{fig:202020-2030-5(5)-555}
\end{figure*}

In our next simulation, we considered coupling of tensors and matrices with some degree of sharing among them. We created a matrix of rank $5$ and a tensor of multilinear rank  $(5,5,5)$  and let them share all $5$ singular components along mode $1$.  Figure \ref{fig:202020-2030-5(5)-555} shows that the coupled norm $\| \cdot \|_{(\mathrm{O},\mathrm{O},\mathrm{O})}^{1}$  had the best performance among the coupled norms for both  matrix and  tensor completion. Individual tensor completion with the overlapped trace norm  had the same performance as  $\| \cdot \|_{(\mathrm{O},\mathrm{O},\mathrm{O})}^{1}$. The NC-Tucker method  performed better than the  coupled norms for tensor  and matrix completion.

\begin{figure*}[h]
 \centering
   \begin{minipage}[h]{0.46\linewidth}
\centering
  {\includegraphics[width=0.99\textwidth]{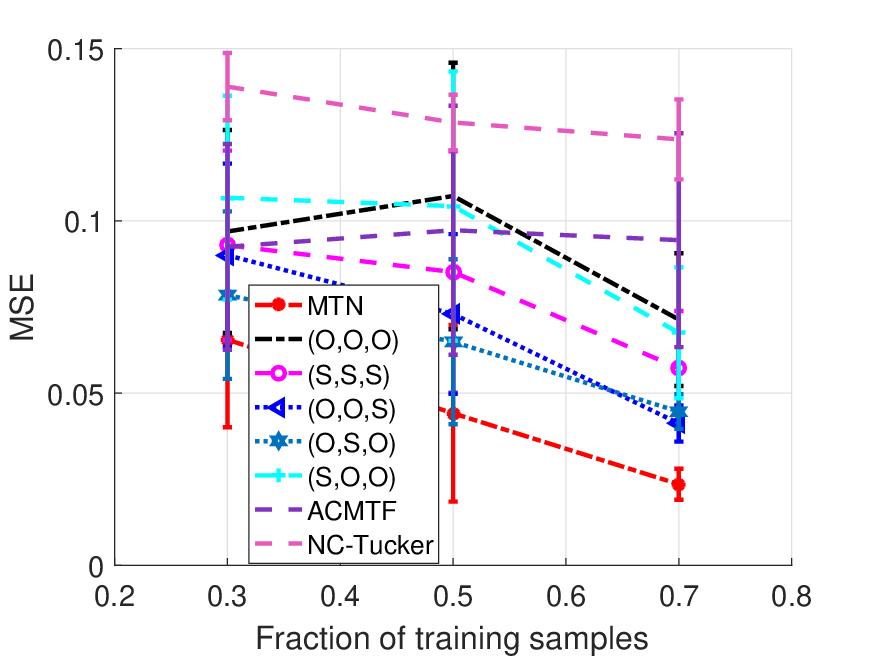}} \\ \vspace{-0.10cm}
(a) Matrix Completion 
\end{minipage}
 \begin{minipage}[h]{0.46\linewidth}
\centering
  {\includegraphics[width=0.99\textwidth]{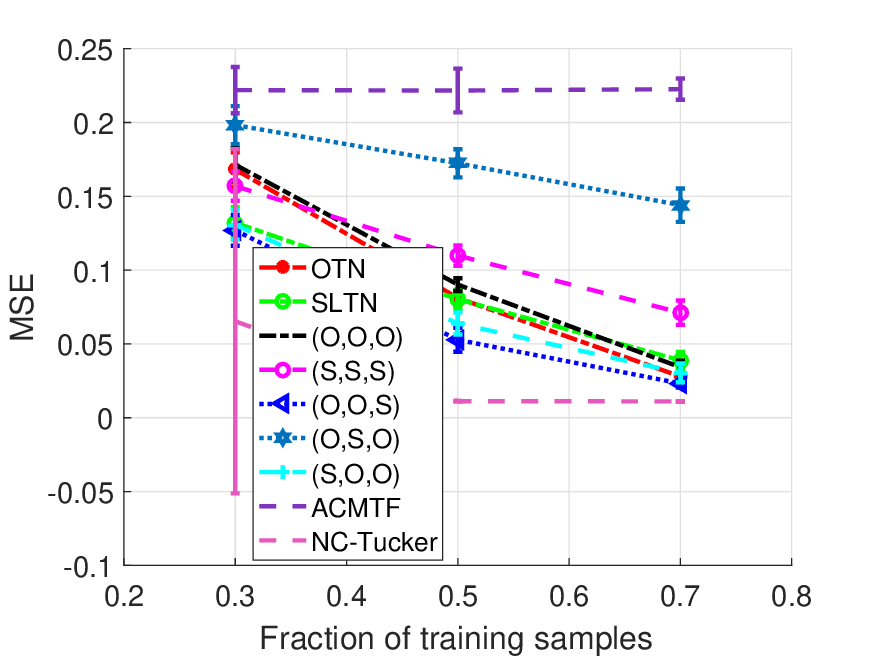}} \\ \vspace{-0.10cm}
(b) Tensor Completion.
\end{minipage}
\caption{Completion performance of matrix  with dimension $20 \times 30$ and rank $5$ and of  tensor with dimension $20 \times 20 \times 20$ and multilinear rank $(5, 15, 5)$ that shared $5$ components.}
\label{fig:202020-2030-5(5)-5155}
\end{figure*}

In our next simulation, we considered a  matrix of rank  $5$ and a tensor of multilinear rank $(5,15,5)$  that  shared all $5$ singular components along mode $1$.  Figure \ref{fig:202020-2030-5(5)-5155} shows that, with this setting, although the coupled norm $\| \cdot \|_{(\mathrm{O},\mathrm{O},\mathrm{S})}^{1}$ had the best performance among the coupled norms and individual tensor completion, it was outperformed by the NC-Tucker method. However, the NC-Tucker method performed poorly in matrix completion compared to the coupled norms. For  the matrix completion, individual matrix completion by the matrix trace norm had  the best performance while  coupled norms $\| \cdot \|_{(\mathrm{O},\mathrm{O},\mathrm{S})}^{1}$ and $\| \cdot \|_{(\mathrm{S},\mathrm{O},\mathrm{O})}^{1}$ had the next best performance.

\begin{figure*}[h!] 
 \centering
 \begin{minipage}[h]{0.45\linewidth}
\centering
  {\includegraphics[width=0.99\textwidth]{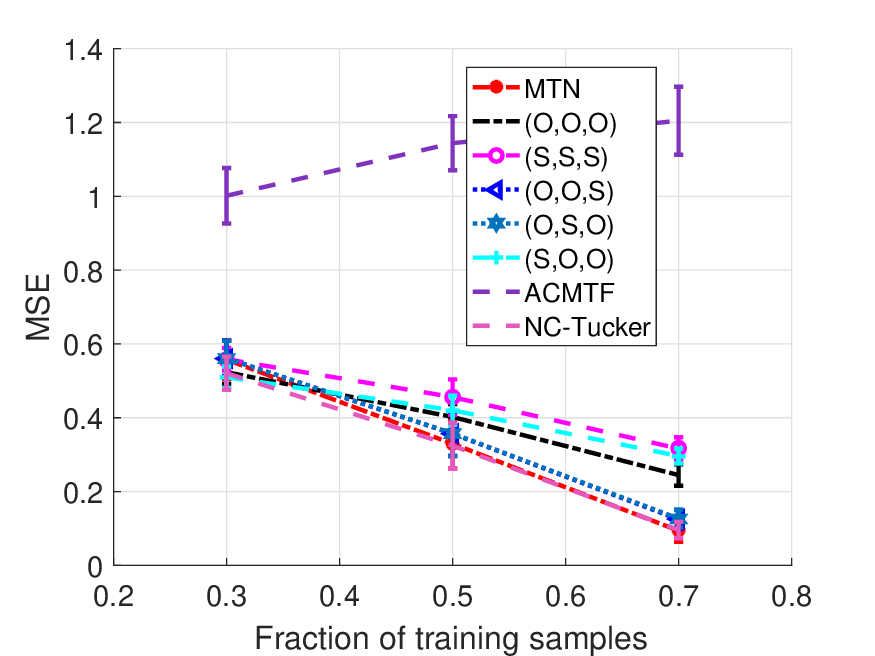}} \\ \vspace{-0.10cm}
(a) Matrix Completion 
\end{minipage}
 \begin{minipage}[h]{0.45\linewidth}
\centering
  {\includegraphics[width=0.99\textwidth]{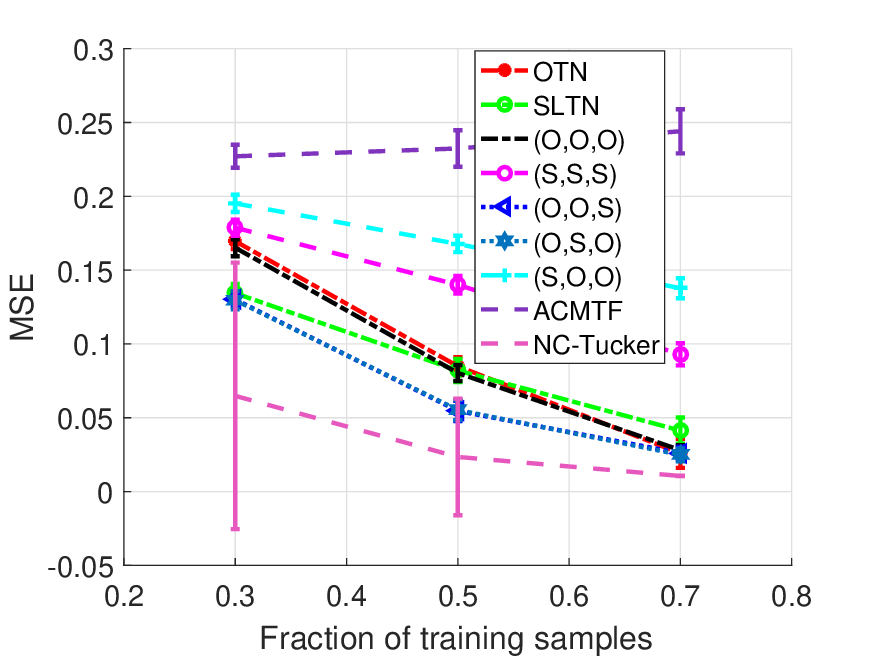}} \\ \vspace{-0.10cm}
(b) Tensor Completion. 
\end{minipage}
\caption{Completion performance of completion of matrix with dimension $20 \times 30$ and rank $5$ and of a tensor with dimension $20 \times 20 \times 20$ and multilinear rank $(15, 5, 5)$ that shared $5$ components.}
  \label{fig:202020-2030-5(5)-1555}
\end{figure*}

For our final simulation, we created a coupled matrix with rank $5$ and a tensor with  multilinear rank $(15,5,5)$, all sharing $5$ singular components along mode $1$.  Figure \ref{fig:202020-2030-5(5)-1555} shows that the mixed coupled norms $\| \cdot \|_{(\mathrm{O},\mathrm{S},\mathrm{O})}^{1}$ and $\| \cdot \|_{(\mathrm{O},\mathrm{O},\mathrm{S})}^{1}$ performed equally and had  better  performance for tensor completion than the individual tensor completion. The NC-Tucker method had  better performance than the coupled norms for tensor completion, while the performance was comparable for matrix completion. For matrix completion when the percentage of training samples was small, coupled norms $\| \cdot \|_{(\mathrm{O},\mathrm{O},\mathrm{O})}^{1}$ and $\| \cdot \|_{(\mathrm{S},\mathrm{O},\mathrm{O})}^{1}$ had better performance. As the  percentage of training samples was increased, the performance of individual matrix completion improved while those of $\| \cdot \|_{(\mathrm{O},\mathrm{S},\mathrm{O})}^{1}$ and $\| \cdot \|_{(\mathrm{O},\mathrm{O},\mathrm{S})}^{1}$ were close but second best.

The results of these simulations show that the ACMTF  performed poorly compared to our proposed methods. 

\subsection{Real-World Data}
As a real-world data experiment, we applied our proposed method to the UCLAF dataset \citep{DBLP:conf/aaai/ZhengCZXY10}, which consists of GPS data for $164$ users in $168$ locations performing $5$ activities, resulting in a sparse user-location-activity tensor $\mathcal{T} \in \mathbb{R}^{164 \times 168 \times 5}$. This dataset also has a user-location matrix $X \in \mathbb{R}^{164 \times 168}$, which we used as side information coupled to the user mode of $\mathcal{T}$. Using  similar observed element  percentages as in the synthetic data simulations we performed completion experiments on $\mathcal{T}$. We considered all the elements of the user-location matrix as observed elements and used them as training data.  We repeated the evaluation  for $10$ random sample selections. We cross-validated the regularization parameters from $0.01$ to $500$ divided into $50$ in  logarithmic linear scale. As a baseline method, we again  used the ACMTF method \citep{DBLP:journals/bmcbi/AcarPGRLNB14} with CP rank $5$. Additionally, we used the coupled (Tucker) method \citep{journals/datamine/ErmisAC15} and the NC-Tucker method with multilinear rank  $(3,3,3)$, where we  selected the best performances among $5$ random initializations.  Figure 10 shows the completion performances for the coupled tensor.

\begin{figure}[h]
  \centering
    \begin{minipage}[h]{0.93\linewidth}
\centering
  {\includegraphics[width=0.99\textwidth]{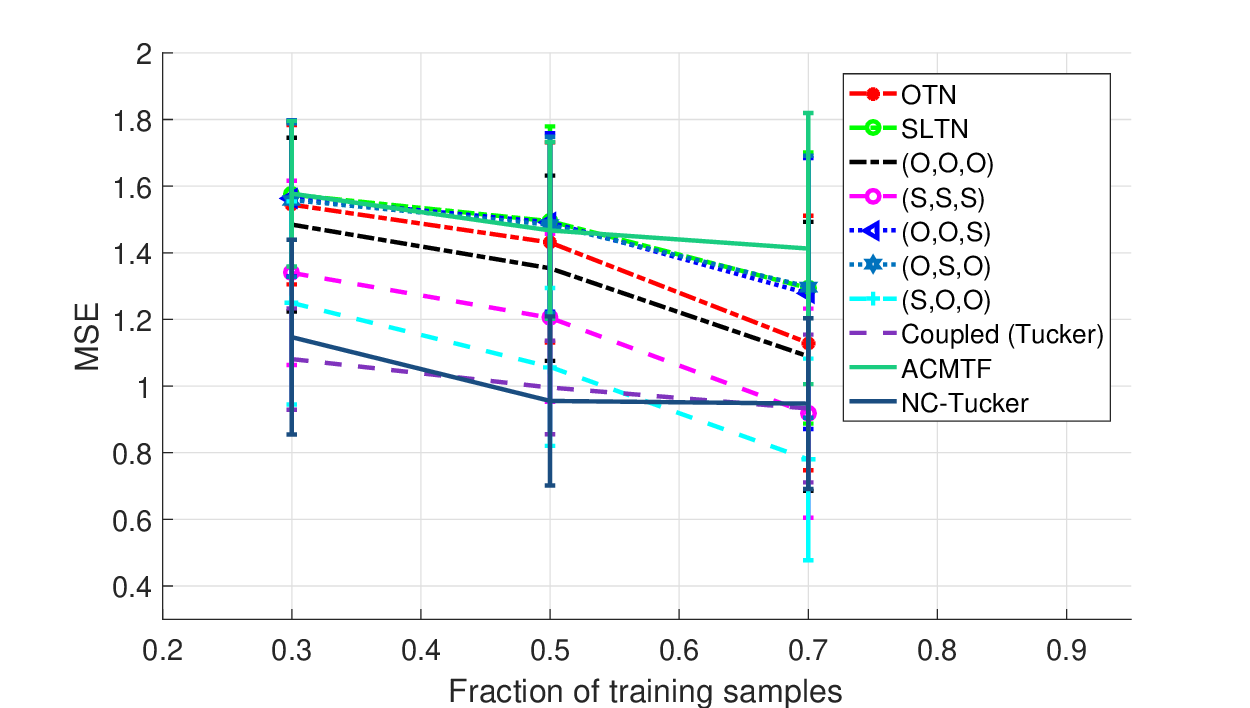} }
Figure 10: Completion performance for UCLAF data. 
\end{minipage}
\end{figure}

We can see that the best performance among coupled norms was that of mixed coupled norm $\| \cdot \|_{(\mathrm{S},\mathrm{O},\mathrm{O})}^{1}$, indicating that learning with side information as a coupled structure improves tensor completion performance compared to completion using only tensor norms. This also indicates that  mode $1$ may have a lower rank than the other modes and that mode $2$ and  $3$ may have ranks closer to each other. The non-convex coupled (Tucker) method and the NC-Tucker method had  better performance than  $\| \cdot \|_{(\mathrm{S},\mathrm{O},\mathrm{O})}^{1}$ when the number of observed samples was less than $70$ percent of the total elements.

\section{Conclusion and Future Work} 
We have proposed a new set of convex norms for the completion problem of coupled  tensors.  We restricted our study to coupling a $3$-way tensor with a matrix and defined low-rank inducing norms by extending trace norms such as the overlapped trace norm and scaled latent trace norm  of tensors and the matrix trace norm. We also introduced the concept of mixed norms, which combines the features  of both overlapped and latent trace norms. We looked at the theoretical properties of our convex completion model and evaluated it using synthetic and real-world data. We found that the proposed coupled norms perform comparably with existing non-convex ones. However,  our norms lead to global optimal solutions and eliminate the need for specifying the ranks of the coupled tensors beforehand. While there are still many aspects to be studied, we believe that our work is the first step in modeling convex norms for coupled  tensors.

Although coupling can occur among many tensors with different dimensions and multiple matrices on different modes, this study focused on a $3$-mode tensor and a single matrix. The methodology used to create coupled norms can be extended to any of those settings, but mere extensions may not lead to the optimal design of norms for those settings. Particularly, the square tensor norm \citep{DBLP:conf/icml/MuHWG14}  has shown to be better suited to tensors beyond three modes and thus can also be used to model novel coupled norms in the future.  Furthermore, theoretical analysis using methods such as the Gaussian width \citep{Amelunxen30062014} may provide  deeper understanding of coupled  tensors, which should enable  design of better norms.  Such studies could be interesting directions for future research.

\section*{Acknowledgment}
MY was supported by the JST PRESTO program JPMJPR165A.
HM has been partially supported by JST ACCEL Grant Number JPMJAC1503
(Japan), MEXT Kakenhi 16H02868 (Japan), FiDiPro by Tekes (currently
Business Finland) and AIPSE programme by Academy of Finland.

\appendix

\section*{Appendices}

\section{Proofs of Dual Norms}
We first provide the proofs of the dual norms of Theorems 1 and 2.

\textit{Proof of Theorem 1.}{
We use  Lemma 3 of  \citep{tomioka/nips13/abs-1303-6370} to prove the duality. Consider a  linear operator $\Phi$ such that $\Phi(\mathcal{T},M) = [ \mathrm{vec}(M); \mathrm{vec}(T_{(1)});\mathrm{vec}(T_{(2)});  \mathrm{vec}(T_{(3)}) ] \in \mathbb{R}^{d_{1}+ 3d_{2}}$, where  $d_{1} = n_{1}m$ and $d_{2} = n_{1}n_{2}n_{3}$. We define  
\begin{equation}
\| z \|_{*} = \Big( \| [Z_{(1)}^{(1)}; X] \|_{S_{p}}^{q} + \sum_{k=2}^{3} \| Z_{(k)}^{(k)} \|_{S_{p}}^{q} \Big)^{1/q}, \label{ed:lm1}
\end{equation}
where $\mathcal{Z}^{(k)}$ is the inverse vectorization of elements $z_{(d_{1}+(k-1)d_{2} + 1):(d_{1}+kd_{2})}$ and $X$  is the inverse vectorization of $z_{1:d_{1}}$. The dual of the above norm is expressed as 
\begin{equation*}
\| z \|_{*^{*}} = \Big( \| [Z_{(1)}^{(1)}; X] \|_{S_{p^{*}}}^{q^{*}} + \sum_{k=2}^{3} \| Z_{(k)}^{(k)}  \|_{S_{p^{*}}}^{q^{*}} \Big)^{1/q^{*}}.
\end{equation*}

Let
\begin{equation*}
\Phi^{\top}(z) = \{\mathcal{T},M\} = \Bigg\{ \sum_{k=1}^{3}\mathcal{Z}^{(k)},X \Bigg\},
\end{equation*}
then  following the Lemma 3  of \citep{tomioka/nips13/abs-1303-6370}, we write
\begin{equation*}
||| [\mathcal{T},M] |||_{\overline{*}(\Phi)} = \inf \| z \| \quad s.t \quad \Phi^{\top}(z) = \{\mathcal{T},M\}.  
\end{equation*}
Given that
\begin{equation*}
||| [\mathcal{T},M] |||_{\underline{*}(\Phi)} := \| [\mathcal{T},M] \|_{(\mathrm{O},\mathrm{O},\mathrm{O}),\underline{S_{p}/q}}^{1},  
\end{equation*}
and following  Lemma 3 in  \citep{tomioka/nips13/abs-1303-6370} we obtain the dual of 
$\| [\mathcal{T},M] \|_{(\mathrm{O},\mathrm{O},\mathrm{O}),\underline{S_{p}/q}}^{1}$ as $\| [\mathcal{T},M] \|_{(\mathrm{L},\mathrm{L},\mathrm{L}),\overline{S_{p^{*}}/q^{*}}}^{1}$.
\QEDB
}

\textit{Proof of Theorem 2.}{
We can apply  Theorem 1 to latent tensors  $\mathcal{T}^{(1)}$ and $\mathcal{T}^{(2)}$ as well as the dual of the overlapping norm to $\mathcal{T}$. First consider the dual with respect to $\mathcal{T}^{(1)}$ and $\mathcal{T}^{(2)}$;  by applying Theorem 1, we obtain
\begin{equation*}
\| \mathcal{T} , M  \|_{(\mathrm{L},\mathrm{O},\mathrm{O}),\overline{S_{p^{*}}/q^{*}}}^{1} =   \Bigg( \Big( \sum_{i}^{r_{1}} \sigma_{i}\big( [T_{(1)};M]\big)^{p^{*}} \Big)^{\frac{q^{*}}{p^{*}}} +  \| \mathcal{T} \|_{(-,\mathrm{O},\mathrm{O}),S_{p}^{*}} \Bigg)^{\frac{1}{q^{*}}}.
\end{equation*} 
Next, by applying  Lemma 1 of \citep{tomioka/nips13/abs-1303-6370} to $\| \mathcal{T} \|_{(-,\mathrm{O},\mathrm{O})}$, we obtain
\begin{multline*}
\| \mathcal{T} , M  \|_{(\mathrm{L},\mathrm{O},\mathrm{O}),\overline{S_{p^{*}}/q^{*}}}^{1} =   \Bigg( \Big( \sum_{i}^{r_{1}} \sigma_{i}\big( [T_{(1)};M]\big)^{p^{*}} \Big)^{\frac{q^{*}}{p^{*}}} \\
 +  \underset{\mathcal{\hat{T}}^{(1)} +  \mathcal{\hat{T}}^{(2)} = \mathcal{T} }{\inf} \Bigg( \Big( \sum_{j}^{r_{2}} \sigma_{j}\big(\hat{T}_{(2)}^{(1)} \big)^{p^{*}} \Big)^{\frac{q^{*}}{p^{*}}} 
  + \Big( \sum_{k}^{r_{3}} \sigma_{k}\big(\hat{T}_{(3)}^{(2)} \big)^{p^{*}} \Big)^{\frac{q^{*}}{p^{*}}} \Bigg) \Bigg)^{\frac{1}{q^{*}}}. 
\end{multline*} 
This completes the proof.
\QEDB
}

\section{Proofs of Excess Risk Bounds}
Here we derive the excess risk bounds for the coupled completion problem.

From previous work \citep{transductive_rademacher,JMLR:v15:shamir14a}, we know that for a loss function  $l(\cdot,\cdot)$ that is a $\Lambda$-Lipschitz loss function and bounded as  $\sup_{i_{1},i_{2},i_{3} } |l(\bold{X}_{i_{1},i_{2},i_{3}}, \bold{W}_{i_{1},i_{2},i_{3}})| \leq b_{l}$ and with the assumption that $|\mathrm{S}_{\mathrm{Train}}| = |\mathrm{S}_{\mathrm{Test}}| = |S|/2$,  we  have the following bound for \eqref{eq:coup_objective_thory} based on  transductive Rademacher complexity theory \citep{transductive_rademacher,JMLR:v15:shamir14a} with probability $1-\delta$, 
\begin{multline*}
 \frac{1}{|\mathrm{S}_{\mathrm{Test}}|} \sum_{(i_{1},i_{2},i_{3}) \in \mathrm{S}_{\mathrm{Test}} } l(\bold{X}_{i_{1},i_{2},i_{3}}, \bold{W}_{i_{1},i_{2},i_{3}}) \\ 
-  \frac{1}{|\mathrm{S}_{\mathrm{Train}}|} \sum_{(i_{1},i_{2},i_{3}) \in \mathrm{S}_{\mathrm{Train}} } l(\bold{X}_{i_{1},i_{2},i_{3}}, \bold{W}_{i_{1},i_{2},i_{3}})  \\ 
\leq  4R(\bold{W}) + b_{l}\bigg(\frac{11 + 4\sqrt{\log{\frac{1}{\delta}}}}{\sqrt{|S_{\mathrm{Train}}|}}\bigg),\label{eq:transductive_formulation1}
\end{multline*} 
where  $R(\bold{W})$ is  transductive Rademacher complexity  defined as
\begin{equation}
R(\bold{W}) = \frac{1}{|\mathrm{S}|}\mathbb{E}_{\sigma} \bigg[  \sup_{\|\bold{W}\|_{\mathrm{cn}} \leq B}  \sum_{(i_{1},i_{2},i_{3}) \in \mathrm{S} }\sigma_{i_{1},i_{2},i_{3}}l(\bold{W}_{i_{1},i_{2},i_{3}} ,\bold{X}_{i_{1},i_{2},i_{3} } ) \bigg] \label{eq:rademacher_complexity2}
\end{equation}
where $\sigma_{i_{1},i_{2},i_{3}} \in \{-1,1\}$ with probability $0.5$ if $(i_{1},i_{2},i_{3}) \in \mathrm{S}$, or $0$ otherwise.

We can rewrite \eqref{eq:rademacher_complexity2} as
\begin{equation*}
\begin{split}
R(\bold{W}) &= \frac{1}{|\mathrm{S}|}\mathbb{E}_{\sigma} \bigg[ \sup_{{\|\bold{W}\|_{\mathrm{cn}} \leq B_{M} + B_{\mathcal{T}}}} \sum_{(i_{1},i_{2},i_{3}) \in \mathrm{S} }   \sigma_{i_{1},i_{2},i_{3} }l(\bold{W}_{i_{1},i_{2},i_{3} } ,\bold{X}_{i_{1},i_{2},i_{3}} ) \bigg]\\ 
&\leq \frac{\Lambda}{|\mathrm{S}|}  \mathbb{E}_{\sigma} \sup_{\|\bold{W}\|_{\mathrm{cn}} \leq B_{M} + B_{\mathcal{T}}} \sum_{(i_{1},i_{2},i_{3})\in \mathrm{S}} \sigma_{i_{1},i_{2},i_{3}} \bold{W}_{i_{1},i_{2},i_{3}} \; \mathrm{(Rademacher\;contraction)} \\
&\leq \frac{\Lambda}{|\mathrm{S}|}  \mathbb{E}_{\sigma}  \sup_{\|\bold{W}\|_{{\mathrm{cn}}} \leq B_{M} + B_{\mathcal{T}}}  \| \bold{W}\|_{{\mathrm{cn}}}  \| \Sigma \|_{\mathrm{cn}^{*}} \; \mathrm{(Holder's\;inequality)}
\end{split},
\end{equation*}
where we have used that $\| \mathcal{W} \|_{\mathrm{F}} \leq B_{\mathcal{T}}$ and $\| W_{M} \|_{\mathrm{F}} \leq B_{M}$, and $\Sigma$ is of dimensions of the coupled tensor consisting  Rademacher variables ($\Sigma_{i_{1},i_{2},i_{3}} = \sigma_{i_{1},i_{2},i_{3}}$ if $(i_{1},i_{2},i_{3}) \in \mathrm{S}$, else $\Sigma_{i_{1},i_{2},i_{3}} =0$).

\textit{Proof of Theorem 3}:{ 
Let $\bold{W} = \mathcal{W} \cup W_{M}$, where $\mathcal{W}$ and $W_{M}$ are the completed tensors of $\mathcal{T}$ and $M$, and let  $\Sigma = \Sigma_{\mathcal{T}} \cup \Sigma_{M}$, where $\Sigma_{\mathcal{T}}$ and $\Sigma_{M}$ consist of the corresponding Rademacher variables ($\sigma_{i_{1},i_{2},i_{3}}$) for  $\mathcal{T}$ and $M$. 
Since we use an overlapping norm, we have $\| \bold{W} \|_{\mathrm{cn}} = \| \mathcal{W},W_{M} \|^{1}_{(\mathrm{O},\mathrm{O},\mathrm{O})}$ from  which we obtain
\begin{equation*}
\begin{split}
\| \mathcal{W},W_{M} \|^{1}_{(\mathrm{O},\mathrm{O},\mathrm{O})} &= \| [W_{(1)};W_{M}] \|_{\mathrm{tr}} + \sum_{k=2}^{3} \| W_{(k)} \|_{\mathrm{tr}} \\
 &\leq   \sqrt{r_{(1)}}(B_{\mathcal{T}} + B_{M}) + \sum_{k=2}^{3}\sqrt{r_{k}}B_{\mathcal{T}},
\end{split}
\end{equation*}
where $(r_{1},r_{2},r_{3})$ is the multilinear rank of $\mathcal{W}$ and $r_{(1)}$ is the rank of the concatenated matrix of unfolding tensors on mode $1$. To obtain the above inequality, we used the fact that, for any matrix $U$ with rank $r$, we have $\| U \|_{\mathrm{tr}} \leq \sqrt{r}\| U \|_{\mathrm{F}}$ \citep{tomioka/nips13/abs-1303-6370}.

Using Lata{\l}a's Theorem \citep{MR2111932,JMLR:v15:shamir14a} for the mode $k$ unfolding, we can bound  $\| \Sigma_{\mathcal{T}{(k)}} \|_{\mathrm{op}}$  
\begin{equation*}
\mathbb{E}\| \Sigma_{\mathcal{T}(k)}  \|_{\mathrm{op}} \leq  C_{1} \Bigg( \sqrt{n_{k}} +  \sqrt{\prod_{j \neq k}^{3}{n_{j}}} +  \sqrt[4]{|\Sigma_{\mathcal{T}(k)}|}\Bigg),
\end{equation*}
and since $\sqrt[4]{|\Sigma_{\mathcal{T}(k)}|} \leq \sqrt[4]{ \prod_{i=1}^{3}{n_{i}}}  \leq \frac{1}{2} \Bigg(\sqrt{n_{k}} +  \sqrt{\prod_{j \neq k}^{3}{n_{j}}} \Bigg)$, we have,
\begin{equation*}
\mathbb{E}\| \Sigma_{\mathcal{T}(k)}  \|_{\mathrm{op}} \leq  \frac{3C_{1}}{2} \Bigg(\sqrt{n_{k}} +  \sqrt{\prod_{j \neq k}^{3}{n_{j}}} \Bigg). 
\end{equation*}
Similarly, using the Lata{\l}a's Theorem, we obtain
\begin{equation*}
\mathbb{E}\| [\Sigma_{\mathcal{T}(1)};\Sigma_M] \|_{\mathrm{op}} \leq  \frac{3C_{2}}{2} \Bigg(\sqrt{n_{1}} + \sqrt{\prod_{j=2}^{3}{n_{j}} + m} \Bigg).
\end{equation*}
To bound  $\mathbb{E}\| \Sigma_{\mathcal{T}}, \Sigma_{M}  \|^{1}_{(\mathrm{O},\mathrm{O},\mathrm{O})^{*}}$, we use the duality relationship from Theorem 1 and Corollary 1
\begin{multline*}
\| \Sigma_{\mathcal{T}}, \Sigma_{M}  \|^{1}_{(\mathrm{O},\mathrm{O},\mathrm{O})^{*}} = \\
 \inf_{ \Sigma_{\mathcal{T}}^{(1)} + \Sigma_{\mathcal{T}}^{(2)} + \Sigma_{\mathcal{T}}^{(3)}  =  \Sigma_{\mathcal{T}}}  \max \Big\{ \| [\Sigma_{{\mathcal{T}(1)}}^{(1)};\Sigma_{M}] \|_{\mathrm{op}}, \| \Sigma_{\mathcal{T}(2)}^{(2)}\|_{\mathrm{op}} , \| \Sigma_{\mathcal{T}(3)}^{(3)}\|_{\mathrm{op}} \Big\}.
\end{multline*}
Since we can take any $\Sigma^{(k)}_{\mathcal{T}}$ to be equal to $\Sigma_{\mathcal{T}}$, the above norm can be upper bounded:
\begin{equation*}
\| \Sigma_{\mathcal{T}}, \Sigma_{M}  \|^{1}_{(\mathrm{O},\mathrm{O},\mathrm{O})^{\star}} \leq \max \Big\{  \| [\Sigma_{\mathcal{T}(1)};\Sigma_{M}] \|_{\mathrm{op}}, \min \big\{ \| \Sigma_{\mathcal{T}(2)} \|_{\mathrm{op}} ,  \| \Sigma_{\mathcal{T}(3)} \|_{\mathrm{op}} \big\} \Big\}. \label{eq:20}
\end{equation*}
Taking the expectation leads to
\begin{equation*}
\begin{split}
\mathbb{E}\| \Sigma_{\mathcal{T}}, \Sigma_{M}  \|^{1}_{(\mathrm{O},\mathrm{O},\mathrm{O})^{*}}  &\leq \mathbb{E}  \max \Big\{  \| [\Sigma_{\mathcal{T}(1)};\Sigma_{M}] \|_{\mathrm{op}}, \min \big\{ \| \Sigma_{\mathcal{T}(2)} \|_{\mathrm{op}} ,  \| \Sigma_{\mathcal{T}(3)} \|_{\mathrm{op}} \big\} \Big\} \\
&\leq \max \Big\{ \mathbb{E} \| [\Sigma_{\mathcal{T}(1)};\Sigma_{M}] \|_{\mathrm{op}}, \min \big\{ \mathbb{E} \| \Sigma_{\mathcal{T}(2)} \|_{\mathrm{op}} , \mathbb{E} \| \Sigma_{\mathcal{T}(3)} \|_{\mathrm{op}} \big\} \Big\}.\label{eq:21}
\end{split}
\end{equation*}
Finally, we have
\begin{equation*}
\begin{split}
R(\bold{W}) &\leq \frac{3\Lambda}{2|\mathrm{S}|} \Big[  \sqrt{r_{(1)}}(B_{\mathcal{T}} + B_{M}) + \sum_{k=2}^{3}\sqrt{r_{k}}B_{\mathcal{T}} \Big]
\\ 
& \qquad\qquad \max \Bigg\{  C_{2}\bigg(\sqrt{n_{1}} + \sqrt{\prod_{j=2}^{3}{n_{j}} + m} \bigg), \min_{k \in {2,3}} C_{1} \bigg( \sqrt{n_{k}} +  \sqrt{\prod_{j \neq k}^{3}{n_{j}}} \bigg)  \Bigg\}.
\end{split}
\end{equation*} \QEDB

Before we give the excess risk bound for the $\| \cdot \|^{1}_{(\mathrm{S},\mathrm{S},\mathrm{S})}$, in the following theorem we give the excess risk of coupled completion with  the $\| \cdot \|^{1}_{(\mathrm{L},\mathrm{L},\mathrm{L})}$.

\begin{thm} 
Let $\|\cdot \|_{\mathrm{cn}} = \| \cdot \|^{1}_{(\mathrm{L},\mathrm{L},\mathrm{L})}$; then, with probability $1-\delta$
\begin{equation*}
\begin{split}
R(\bold{W}) &\leq \frac{3\Lambda}{2|\mathrm{S}|} \Bigg[ \sqrt{r_{(1)}}B_{M}
 + \min \Big(\sqrt{r_{(1)}}, \min_{k=2,3}\sqrt{r_{k}} \Big)B_{\mathcal{T}} \Bigg]\\
 & \qquad \max \Bigg\{C_{2} \Bigg(\sqrt{n_{1}} + \sqrt{\prod_{j=2}^{3}{n_{j}} + m} \Bigg),  \max_{k=2,3} \bigg\{ C_{2}\Bigg(\sqrt{n_{k}} +  \sqrt{\prod_{j \neq k}^{3}{n_{j}}} \Bigg) \bigg\} \Bigg\},
\end{split}
\end{equation*}
where $(r_{1},r_{2},r_{3})$ is the multilinear rank of $\mathcal{W}$, $r_{(1)}$ is the rank of the coupled unfolding on mode $1$ and $B_{M}$, $B_{\mathcal{T}}$, $C_{1}$, and $C_{2}$ are constants.
\end{thm}
\textit{Proof}:{ 
Again, let $\bold{W} = \mathcal{W} \cup W_{M}$, where $\mathcal{W}$ and $W_{M}$ are the completed tensors of $\mathcal{T}$ and $M$, and $\Sigma = \Sigma_{\mathcal{T}} \cup \Sigma_{M}$, where $\Sigma_{\mathcal{T}}$ and $\Sigma_{M}$ consist of the corresponding Rademacher variables. 
We can see that
\begin{equation*}
\| \bold{W} \|^{1}_{(\mathrm{L},\mathrm{L},\mathrm{L})} = \inf_{\mathcal{W}^{(1)} + \mathcal{W}^{(2)} + \mathcal{W}^{(3)} = \mathcal{W}}{\Big( \|[W_{(1)}^{(1)};W_{M}]\|_{\mathrm{tr}} + \sum_{k=2}^{3} \| W_{(k)}^{(k)}\|_{\mathrm{tr}} \Big)}, 
\end{equation*}
which can be bounded as 
\begin{equation*}
\| \bold{W} \|^{1}_{(\mathrm{L},\mathrm{L},\mathrm{L})} \leq \sqrt{r_{(1)}}(B_{M} + B_{\mathcal{T}}) + \min_{k=2,3}\sqrt{r_{k}}B_{\mathcal{T}},
\end{equation*}
where the last term is derived by considering the infimum  with respect to $\mathcal{W}^{(2)}$ and $\mathcal{W}^{(3)}$.

Using  the duality result given in Theorem 1 (Corollary 1) and Lata{\l}a's Theorem, we obtain
\begin{equation*}
\begin{split}
\| \Sigma_{\mathcal{T}}, \Sigma_{M} \|^{1}_{(\mathrm{L},\mathrm{L},\mathrm{L})^{*}} &\leq \max \Big\{ \mathbb{E}\| [\Sigma_{\mathcal{T}(1)}; \Sigma_{M}]\|_{\mathrm{op}}, \mathbb{E}\| \Sigma_{\mathcal{T}(2)}\|_{\mathrm{op}}, \mathbb{E}\| \Sigma_{\mathcal{T}(3)}\|_{\mathrm{op}} \Big\} \\
 &\leq \frac{3}{2} \max \Bigg\{C_{2} \Bigg(\sqrt{n_{1}} + \sqrt{\prod_{j=2}^{3}{n_{j}} + m} \Bigg) , \\
 & \qquad \qquad \qquad \qquad \qquad \max_{k=2,3} \bigg\{ C_{1}\Bigg(\sqrt{n_{k}} +  \sqrt{\prod_{j \neq k}^{3}{n_{j}}} \Bigg) \bigg\} \Bigg\}.
\end{split}
\end{equation*}
Finally, we have
\begin{equation*}
\begin{split}
R(\bold{W}) &\leq \frac{3\Lambda}{2|\mathrm{S}|} \Bigg[  \sqrt{r_{(1)}}(B_{M} + B_{\mathcal{T}}) + \min_{k=2,3}\sqrt{r_{k}}B_{\mathcal{T}} \Bigg]\\
 & \qquad \max \Bigg\{C_{2} \Bigg(\sqrt{n_{1}} + \sqrt{\prod_{j=2}^{3}{n_{j}} + m} \Bigg) ,  \max_{k=2,3} \bigg\{ C_{1}\Bigg(\sqrt{n_{k}} +  \sqrt{\prod_{j \neq k}^{3}{n_{j}}} \Bigg) \bigg\} \Bigg\} .
\end{split}
\end{equation*}
\QEDB
}

\textit{Proof of Theorem 4}:{ 
By definition, we have
\begin{equation*}
\| \bold{W} \|^{1}_{(\mathrm{S},\mathrm{S},\mathrm{S})} = \inf_{\mathcal{W}^{(1)} + \mathcal{W}^{(2)} + \mathcal{W}^{(3)} = \mathcal{W}}{\Big( \frac{1}{\sqrt{n_{1}}}\|[W_{(1)}^{(1)},W_{M}]\|_{\mathrm{tr}} + \sum_{k=2,3} \frac{1}{\sqrt{n_{k}}} \| W_{(k)}^{(k)}\|_{\mathrm{tr}} \Big)},
\end{equation*}
which results in
\begin{equation*}
\| \bold{W} \|^{1}_{(\mathrm{S},\mathrm{S},\mathrm{S})} \leq \sqrt{\frac{r_{(1)}}{n_{1}}}(B_{M} + B_{\mathcal{T}}) + \min_{k \in 2,3} \sqrt{\frac{r_{k}}{n_{k}}}B_{\mathcal{T}}.
\end{equation*}

Using  the duality result given in Theorem 1 and  Lata{\l}a's Theorem, we obtain
\begin{equation*}
\begin{split}
\mathbb{E}\| \Sigma_{\mathcal{T}}, \Sigma_{M} \|^{1}_{(\mathrm{S},\mathrm{S},\mathrm{S})^{*}} &= \mathbb{E}\max \bigg\{ \sqrt{n_{1}}\| [\Sigma_{\mathcal{T}(1)};\Sigma_{M}]\|_{\mathrm{op}}, \sqrt{n_{2}}\| \Sigma_{\mathcal{T}(2)}\|_{\mathrm{op}} , \sqrt{n_{3}} \| \Sigma_{\mathcal{T}(3)}\|_{\mathrm{op}} \bigg\} \\
 &\leq \frac{3}{2} \max \Bigg\{ C_{2}\Bigg(n_{1} + \sqrt{\prod_{i=1}^{3}{n_{i}} + n_{1}m }\Bigg),   C_{1} \max_{k=2,3}\Bigg(n_{k} + \sqrt{\prod_{i \neq k}^{3}{n_{i}}}\Bigg) \Bigg\}.
\end{split}
\end{equation*}
Finally, we have
\begin{equation*}
\begin{split}
R(\bold{W}) &\leq \frac{3\Lambda}{2|\mathrm{S}|} \Bigg[ \sqrt{\frac{r_{(1)}}{n_{1}}}(B_{M} + B_{\mathcal{T}}) + \min_{k \in 2,3} \sqrt{\frac{r_{k}}{n_{k}}}B_{\mathcal{T}} \Bigg]\\
 & \qquad \max \Bigg\{ C_{2}\Bigg(n_{1} + \sqrt{\prod_{i=1}^{3}{n_{i}} + n_{1}m }\Bigg) , C_{1} \max_{k=2,3}\Bigg(n_{k} + \sqrt{\prod_{i = 1}^{3}{n_{i}}}\Bigg) \Bigg\}.
\end{split}
\end{equation*}
\QEDB
}

\textit{Proof of Theorem 5}:{
First let us look at  $\| \bold{W} \|^{1}_{(\mathrm{S},\mathrm{O},\mathrm{O})}$, which is expressed as
\begin{equation*}
\| \bold{W} \|^{1}_{(\mathrm{S},\mathrm{O},\mathrm{O})} = \underset{\mathcal{W}^{(1)} + \mathcal{W}^{(2)} = \mathcal{W} }{\inf} \Big( \frac{1}{\sqrt{n_{1}}} \| [W_{(1)}^{(1)};W_{M}] \|_{\mathrm{tr}} 
 + \|W_{(2)}^{(2)} \|_{\mathrm{tr}}   +  \|W_{(3)}^{(2)} \|_{\mathrm{tr}} \Big).
 \end{equation*}
This norm can be upper bounded 
\begin{equation*}
\| \bold{W} \|^{1}_{(\mathrm{S},\mathrm{O},\mathrm{O})} \leq \sqrt{\frac{r_{(1)}}{n_{1}}}(B_{M} + B_{\mathcal{T}})  +  \sum_{i=2,3} \sqrt{r_{i}}  B_{\mathcal{T}}.
\end{equation*}
Now we are left with bounding $\| \Sigma_{\mathcal{T}}, \Sigma_{M} \|^{1}_{(\mathrm{S},\mathrm{O},\mathrm{O})^{*}}$. Using  Theorem 2, we obtain
\begin{equation*}
\| \Sigma_{\mathcal{T}}, \Sigma_{M} \|_{(\mathrm{S},\mathrm{O},\mathrm{O})^{*}}^{1} \leq \max \Big(\sqrt{n_{1}}\| [\Sigma_{\mathcal{T}(1)}; \Sigma_{M}] \|_{\mathrm{op}}, \min \big(\| \Sigma_{\mathcal{T}(2)} \|_{\mathrm{op}}, \| \Sigma_{\mathcal{T}(3)} \|_{\mathrm{op}} \big)  \Big),
\end{equation*}
We then have 
\begin{equation*}
\mathbb{E}  \| \Sigma_{\mathcal{T}}, \Sigma_{M} \|_{(\mathrm{S},\mathrm{O},\mathrm{O})^{*}}^{1} \leq \frac{3}{2}\max \Bigg\{ C_{2}\Bigg( n_{1} + \sqrt{\prod_{i=1}^{3}n_{i} + n_{1}m} \Bigg),  \min_{k=2,3} C_{1} \Bigg(\sqrt{n_{k}} + \sqrt{\prod_{i \neq k}^{3}n_{i}}\Bigg) \Bigg\},
\end{equation*}
The final resulting bound is 
\begin{equation*}
\begin{split}
R(\bold{W}) &\leq \frac{3\Lambda}{2|\mathrm{S}|} \Bigg[\sqrt{\frac{r_{(1)}}{n_{1}}}(B_{M} + B_{\mathcal{T}})  +  \sum_{i=2,3} \sqrt{r_{i}}  B_{\mathcal{T}}\Bigg] \\
 & \qquad \max  \Bigg\{ C_{2}\Bigg( n_{1} + \sqrt{\prod_{i=1}^{3}n_{i} + n_{1}m} \Bigg),  \min_{k=2,3} C_{1} \Bigg(\sqrt{n_{k}} + \sqrt{\prod_{i \neq k}^{3}n_{i}}\Bigg) \Bigg\}.
\end{split}
\end{equation*}
\QEDB
}

In addition to the above transductive bounds for completion with coupled  norms, we also provide the bounds for individual tensor completion with tensor norms such as the overlapped trace norm, the latent trace norm, and the scaled latent trace norm. We can consider \eqref{eq:coup_objective_thory} only for a matrix or a tensor without coupling and with low-rank regularization. Therefore, we may have the transductive bounds for a matrix $M \in \mathbb{R}^{n_{1} \times m}$ \citep{JMLR:v15:shamir14a} as    
\begin{equation}
R(W_{M})  \leq c \frac{B_{M} \Lambda}{|\mathrm{S}^{M}|} \sqrt{\hat{r}}\bigg(\sqrt{n_{1}} +  \sqrt{m} \bigg),
\end{equation}
where $\mathrm{S}^{M}$ is the index set of observed samples of  matrix $M$, $\hat{r}$ is the rank induced by  matrix trace norm regularization, and $c$ is a constant.

Next we can consider  the transductive bounds for tensor $\mathcal{T} \in \mathbb{R}^{n_{1} \times n_{2} \times n_{3}}$ with regularization using norms such as the overlapped trace norm \citep{tomioka/nips13/abs-1303-6370}, the latent trace norm \citep{tomioka/nips13/abs-1303-6370} and the scaled latent trace norm \citep{nips-14} in the following three theorems. We denote the index set of observed sample of $\mathcal{T}$ by $\mathrm{S}^{\mathcal{T}}$.

\begin{thm} Using the overlapped trace norm regularization given as $\|\mathcal{W}\|_{\mathrm{overlap}} = \|\mathcal{W}\|_{(\mathrm{O},\mathrm{O},\mathrm{O})} $, we obtain 
\begin{equation*}
R(\mathcal{W})  \leq c_{1}  \frac{B_{\mathcal{T}} \Lambda}{|\mathrm{S}^{\mathcal{T}}|} \bigg(\sum_{k=1}^{3} \sqrt{\hat{r}_{k}}\bigg)\min_{k}\bigg(\sqrt{n_{k}} +  \sqrt{ \prod_{j \neq k}^{3}n_{j}} \bigg), 
\end{equation*}
for some constant $c_{1}$; $(\hat{r}_{1},\hat{r}_{2},\hat{r}_{3})$ is the multilinear rank of $\mathcal{W}$. 
\end{thm}

\textit{Proof}:
{Using the same procedure as for Theorem 3, we obtain 
\begin{equation*}
\mathbb{E}\|\Sigma_{\mathcal{T}} \|_{{\mathrm{overlap}}^{*}} \leq \mathbb{E}\min_{k}  \|\Sigma_{\mathcal{T}(k)}\|_{\mathrm{op}} \leq \min_{k} \mathbb{E} \|\Sigma_{\mathcal{T}(k)}\|_{\mathrm{op}} \leq  \frac{3c_{1}}{2}\min_{k} \bigg(\sqrt{n_{k}} +  \sqrt{ \prod_{j \neq k}^{3}n_{j}} \bigg).
\end{equation*}

Since $\| \mathcal{W}\|_{\mathrm{overlap}} \leq \bigg(\sum_{k=1}^{3} \sqrt{\hat{r}_{k}}\bigg)B_{\mathcal{T}}$, where $\| \mathcal{W}\|_{\mathrm{F}} \leq B_{\mathcal{T}}$ \citep{tomioka/nips13/abs-1303-6370}, we have
\begin{equation*}
R(\mathcal{W})  \leq c_{1}  \frac{B_{\mathcal{T}}  \Lambda}{|\mathrm{S}^{\mathcal{T}}|} \bigg(\sum_{k=1}^{3} \sqrt{\hat{r}_{k}}\bigg)\min_{k} \bigg(\sqrt{n_{k}} + \sqrt{ \prod_{j \neq k}^{3}n_{j}} \bigg). 
\end{equation*} 
\QEDB
}

\begin{thm} Using the latent trace norm regularization given by $\|\mathcal{W}\|_{\mathrm{latent}} = \| \mathcal{W}\|_{(\mathrm{L},\mathrm{L},\mathrm{L})}$, we obtain 
\begin{equation*}
R(\mathcal{W})  \leq c_{2} \Lambda B_{\mathcal{T}}  \frac{\min_{k}\sqrt{\hat{r}_{k}}}{|\mathrm{S}^{\mathcal{T}}|} \max_{k}\bigg(\sqrt{n_{k}} + \sqrt{ \prod_{j \neq k}^{3}n_{j}}\bigg), 
\end{equation*}
for some constant $c_{2}$;   $(\hat{r}_{1},\hat{r}_{2},\hat{r}_{3})$ is the multilinear rank of $\mathcal{W}$.
\end{thm}

\textit{Proof} :{Using  the duality result from \citep{nips-14}, we have
\begin{equation*}
\| \Sigma_{\mathcal{T}} \|_{{\mathrm{latent}}^{*}} = \max_{k} \| \Sigma_{\mathcal{T}(k)} \|_{\mathrm{op}}.\label{eq:22}
\end{equation*}

Using  Lata{\l}a's Theorem, we obtain 
\begin{equation*}
\mathbb{E}\|\Sigma_{\mathcal{T}} \|_{{\mathrm{latent}}^{*}} \leq  \frac{3c_{2}}{2}\max_{k} \bigg(\sqrt{n_{k}} +  \sqrt{ \prod_{j \neq k}^{3}n_{j}} \bigg).
\end{equation*}

Finally, using the known bound $\|\mathcal{W}\|_{\mathrm{latent}} \leq \min_{i}\sqrt{\hat{r}_{i}} B_{\mathcal{T}}$ \citep{nips-14}, where $\| \mathcal{W}\|_{\mathrm{F}} \leq B_{\mathcal{T}}$, we obtain the excess risk:
\begin{equation*}
R(\mathcal{W})  \leq  \frac{3c_{2}\Lambda B_{\mathcal{T}}\min_{i}\sqrt{\hat{r}_{i}}}{2|\mathrm{S}^{\mathcal{T}}|} \max_{k}\bigg(\sqrt{n_{k}} +  \sqrt{ \prod_{j \neq k}^{3}n_{j}} \bigg). 
\end{equation*} 
\QEDB
}

\begin{thm} Using the scaled latent trace norm regularization given by $\|\mathcal{W}\|_{\mathrm{scaled}} = \|\mathcal{W}\|_{(\mathrm{S},\mathrm{S},\mathrm{S})}$, we obtain 
\begin{equation*}
R(\mathcal{W}) \leq   \frac{3c_{3} \Lambda B_{\mathcal{T}}}{2|\mathrm{S}^{\mathcal{T}}|} \min_{i} \bigg(\sqrt{\frac{\hat{r}_{i}}{n_{i}}} \bigg) \max_{k}\bigg(n_{k} + \sqrt{ \prod_{j = 1 }^{3}n_{j}} \bigg). 
\end{equation*}
for some constant $c_{3}$;  $(\hat{r}_{1},\hat{r}_{2},\hat{r}_{3})$ is the multilinear rank of $\mathcal{W}$.
\end{thm}

\textit{Proof}:
{From previous work \citep{nips-14}, we can derive
\begin{equation*}
\| \Sigma_{\mathcal{T}} \|_{{\mathrm{scaled}}^{*}} = \max_{k} \sqrt{n_{k}}  \| \Sigma_{\mathcal{T}(k)} \|_{\mathrm{op}}. \label{eq:23}
\end{equation*} 
Using an approach similar to that for Theorem 8  with the additional scaling of $\sqrt{n_{k}}$ and using  the Lata{\l}a's Theorem, we arrive at the following bound:
\begin{equation*}
R(\mathcal{W}) \leq   \frac{3c_{3} \Lambda B_{\mathcal{T}}}{2|\mathrm{S}^{\mathcal{T}}|} \min_{i} \bigg(\sqrt{\frac{\hat{r}_{i}}{n_{i}}} \bigg) \max_{k}\bigg(n_{k} + \sqrt{ \prod_{j = 1 }^{3}n_{j}} \bigg). 
\end{equation*} 
\qed
}

\end{document}